\pdfoutput=1
\documentclass{article}

\usepackage{PRIMEarxiv}
\usepackage{subcaption}
\usepackage[utf8]{inputenc}
\usepackage[T1]{fontenc}
\usepackage{microtype}
\usepackage{graphicx}
\usepackage{float}
\usepackage{algorithm}
\usepackage{algpseudocode}
\usepackage{multicol}
\usepackage{multirow}
\usepackage{enumitem}
\usepackage{booktabs}
\usepackage{tabularx}
\usepackage{amsmath}
\usepackage{amsfonts}
\usepackage{amssymb}
\usepackage{textcomp}
\usepackage{nicefrac}
\usepackage{tikz}
\usepackage{makecell}
\usepackage{caption}
\usepackage{bbding}
\usepackage[table]{xcolor}
\usepackage{diagbox}
\usepackage{etoc}
\usepackage[numbers,sort&compress]{natbib}
\usepackage{url}
\usepackage{hyperref}
\usepackage{array}

\hypersetup{
  colorlinks=true,
  linkcolor=blue,
  citecolor=blue,
  urlcolor=blue,
  pdfauthor={Yongqi Yu and Yu Zhang},
  pdftitle={TestHallVQA: Exploring LVLMs' Document-Level Reasoning under Redundant Contexts from Scientific Exams}
}

\graphicspath{{media/}}

\definecolor{burgundy}{RGB}{128,0,32}

\newcommand{\circled}[1]{%
  \tikz[baseline=(char.base)]{%
    \node[shape=circle,draw,inner sep=1pt](char){#1};%
  }%
}

\title{TestHallVQA: Exploring LVLMs' Document-Level Reasoning under Redundant Contexts from Scientific Exams}

\author{
  Yongqi Yu\\
  Harbin Institute of Technology\\
  Harbin, China\\
  \texttt{yqyu@ir.hit.edu.cn}\\
  \href{https://orcid.org/0009-0005-0674-788X}{ORCID: 0009-0005-0674-788X}
  \And
  Yu Zhang\thanks{Corresponding author.}\\
  Harbin Institute of Technology\\
  Harbin, China\\
  \texttt{zhangyu@ir.hit.edu.cn}\\
  \href{https://orcid.org/0000-0003-3090-7431}{ORCID: 0000-0003-3090-7431}
}

\date{}

\begin{document}
\maketitle
\thispagestyle{empty}

\begin{abstract}
Large Vision--Language Models (LVLMs) are increasingly expected to perform visual question answering (VQA) over planar media. However, existing planar VQA benchmarks typically emphasize isolated challenges: some emphasize long-document understanding with limited reasoning depth, while others require complex visual reasoning but remain restricted to single-page, noise-free settings. Moreover, through theoretical analysis, we identify the impact of irrelevant visual tokens, which leads to measurable performance degradation but has received little attention with respect to systematic quantification.
To address these limitations, we introduce TestHallVQA, a multi-image VQA benchmark that simultaneously embodies document-level scale and the difficulty of human examinations, while providing comprehensive task coverage. Leveraging TestHallVQA's ability to controllably inject multi-level contextual redundancy, we further propose a novel metric, F1-R\textsuperscript{2}, which jointly quantifies LVLMs' computational reasoning capability and their evidence retrieval robustness against document-level redundancy.
Extensive experiments and analyses on mainstream LVLMs reveal their latent deficiencies across multiple dimensions, offering concrete insights and directions for future research. The associated datasets, code, and complete theoretical derivations are available at \url{https://github.com/yqyu2317/TestHallVQA-benchmark}.
\end{abstract}

\keywords{Document Visual Question Answering \and Multimodal Reasoning \and Large Vision-Language Models}

\section{Introduction}
\label{sec:intro}

With the maturation of cross-modal alignment techniques, large vision-language models (LVLMs) have emerged as a dominant paradigm for extending large language models (LLMs)~\cite{LLM} with visual perception capabilities for multimodal tasks. Meanwhile, planar media---including books, academic papers, and reports---have long served as fundamental carriers of human knowledge dissemination~\cite{eee}. Accordingly, enabling LVLMs to process planar visual content efficiently has become an important research direction \cite{docvqa,longdocurl,dude,mmmu,mathvista}, aiming to reduce the repetitive manual effort required for document understanding and management.

Unlike natural scene images, planar data constitute a prototypical form of \textbf{multimodal media}, in which textual and visual elements---including illustrations, charts, spatial layouts, and other structural components---are tightly intertwined. This intrinsic heterogeneity places substantial demands on LVLMs' fine-grained visual perception and textual understanding.

\begin{figure}
    \centering
    \includegraphics[width=\linewidth]{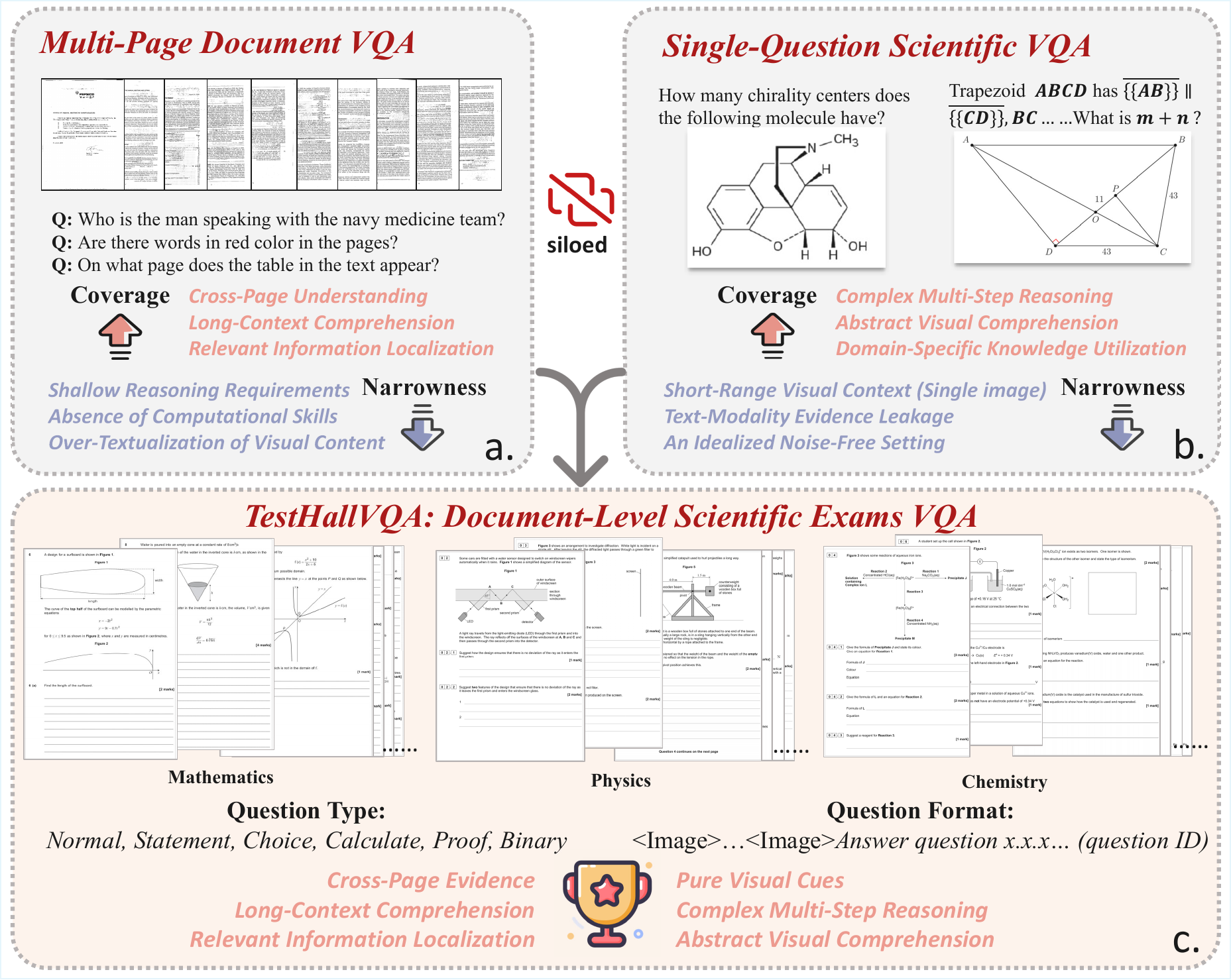}
    \caption{Illustration of the siloed status of existing planar-media VQA benchmarks. TestHallVQA unifies the strengths of both sides, providing a comprehensive challenge.}
    \label{fig:intro}
\end{figure}

Current planar visual question answering (VQA) benchmarks can be broadly categorized into two groups:

\textbf{(1) Multi-Page Document VQA:} These datasets are constructed from text-centric sources, including documents \cite{dude,MP,mmlongbench,longdocurl}, academic papers \cite{paperdataset}, and web pages \cite{visualmrc}. As illustrated in Figure~\ref{fig:intro}a, they pose dual challenges in contextual scale and layout complexity, requiring cross-page and cross-element interactions. However, they remain limited in capturing the depth of complex reasoning. Consequently, they primarily evaluate a model's \emph{retrieval} capability.

\textbf{(2) Single-Question Scientific VQA:} In contrast, this category focuses on highly abstract and challenging planar visuals, such as geometric diagrams and chemical structures \cite{mmmu,mathvista,medicalqa}. As illustrated in Figure~\ref{fig:intro}b, each sample typically consists of a single small image in which the relevant evidence is presented explicitly. Without the information density and contextual complexity of document-level inputs, these samples constitute a self-contained and relatively noise-free setting. Consequently, such benchmarks evaluate only a model's \emph{reasoning} capability.

However, in real-world deployments, critical information is often buried in long and highly redundant contexts. For example, complex reasoning over dozens of pages of documents is a common practical requirement. \textbf{Models cannot know \emph{a priori} which parts of the context will be relevant, and users likewise cannot feasibly identify all supporting evidence manually.} Existing benchmarks, however, either lack sufficient reasoning depth or assume an overly clean context. These two settings are largely orthogonal, each capturing only a limited aspect of real-world complexity.

On the other hand, we theoretically analyze the effect of redundant visual tokens on information flow in LVLM decoders, showing that such redundant tokens contaminate relevant tokens with an exponentially increasing effect across layers. We term this phenomenon \textbf{redundant visual contamination}.

Motivated by the foregoing practical and theoretical considerations, we seek a data source that simultaneously embodies broad contextual scope and substantial reasoning depth, and identify \emph{illustrated examination papers} as an ideal candidate. To this end, we curate a large collection of exam papers and develop a semi-automated annotation pipeline to construct a document-level scientific examination benchmark comprising over 10,000 question--answer pairs and 7,000 paper images. We term this dataset \textbf{TestHallVQA} (Figure~\ref{fig:intro}c), as it places LVLMs in an environment analogous to a real-world examination hall, targets more realistic evaluation settings, and introduces challenges for document-level deep reasoning. Furthermore, to investigate how LVLM performance degrades with increasing redundant context, we equip TestHallVQA with a controlled mechanism for injecting multi-level, highly confounding redundant image pages, thereby inducing smoothly varying contextual \emph{regimes} for each question.

Building on this design, we introduce \textbf{F1-R\textsuperscript{2}} (\textbf{F1} score on \textbf{R}etrieval--\textbf{R}easoning), a unified evaluation metric that jointly integrates the assessment of visual reasoning ability and robustness to redundant visual distractions, rather than decoupling them as in conventional benchmarks. The contributions of this work are threefold:

\begin{itemize}[leftmargin=1em,labelsep=0.5em,itemsep=0pt, topsep=3pt]
    \item We provide preliminary theoretical evidence for the existence of \textbf{``redundant visual contamination''} in LVLMs---a phenomenon that is not captured or diagnosed by mainstream VQA benchmarks.
    \item We introduce \textbf{TestHallVQA}, a multi-image VQA benchmark that not only poses substantial visual and computational challenges, but also supports the controlled injection of document-level visual redundancy, thereby more faithfully reflecting real-world application scenarios.
    \item We propose a novel metric, \textbf{F1-R\textsuperscript{2}}, which jointly quantifies LVLMs' problem-solving capability (Reasoning) and their robustness to redundant visual contamination (Retrieval), moving beyond single-dimensional evaluation.
\end{itemize}

\section{Related Work}

Existing VQA tasks can be broadly grouped into real-world scenes, 3D simulated scenes, and planar-media scenes. The first two, as projections or simulations of the physical world, have abundant annotations and are well-established in prior work \cite{VQA,ok,vizwiz,TDIUC,scanqa}.

Beyond real scenes, planar-media VQA has gained increasing attention. Originating from DocVQA \cite{2021icdar,document}, exemplified by SingleDocVQA \cite{docvqa}, this line introduced document-page datasets as a new planar medium. VisualMRC \cite{visualmrc} scaled this direction with web-based images, while InfographicsVQA \cite{infographicvqa} and ChartQA \cite{chartqa} expanded to infographics and chart images requiring structured visual-semantic reasoning, yet early work was limited to single-page contexts until DUDE \cite{dude} and MP-DocVQA \cite{MP} introduced multi-page inputs demanding cross-page reasoning. MMLongBench \cite{mmlongbench} further extended the context length (average 47.5 pages), while LongDocURL \cite{longdocurl} advanced this to approximately 85 pages with denser cross-element dependencies. Nevertheless, they remain text-dominant, emphasizing information redundancy over genuine reasoning depth, with most answers attainable through shallow inference rather than true chain-of-thought (CoT) \cite{cot} reasoning.

Another line of planar-media VQA centers on deep reasoning. Early studies addressed purely textual corpora (e.g., MATH \cite{MATH}, GSM8K \cite{GSM8K}), later extending to visually grounded reasoning. GeoQA \cite{geoqa}, Geometry3K \cite{Geometry3K}, and UniGeo \cite{unigeo} targeted geometric problem solving; MathVista \cite{mathvista} broadened scope to multimodal mathematics. \citet{mathverse,HC-M3D} identified redundancy between textual and visual content. MATH-Vision \cite{MATH-Vision} enriched task and domain diversity, while WeMath \cite{wemath} introduced process-level evaluation. Other mathematical datasets \cite{MM-math,mathscape,zhang2023m3exam,exams} exhibit complementary strengths. Beyond mathematics, LogicVista \cite{logicvista} examined logical inference over varied planar imagery, and MMMU \cite{mmmu}, MMMU-Pro \cite{mmmu-pro}, SceMQA \cite{scemqa} expanded to broader scientific reasoning. Despite their diversity, these benchmarks assume a noise-free setting with one question per image and no redundant or distracting information, which differs substantially from real-world redundant and complex scenarios.

\section{TestHallVQA Benchmark}

In this section, we provide a comprehensive overview of the proposed TestHallVQA benchmark.

\subsection{Theoretical Motivation: The Phenomenon of Redundant Visual Contamination}
\label{sec:theo}
Due to attention interactions during autoregressive decoding in LVLMs, the introduction of redundant images causes irrelevant token contributions to be mixed into the hidden states of relevant tokens, thereby leading to representation contamination. To characterize how redundant information perturbs the reasoning process of LVLMs, we employ \textbf{a symbolization-based approximate algorithm} to provide a preliminary theoretical investigation of its overall trend. Here, we focus exclusively on the most common self-attention decoder architecture as our analytical model; the derivations for other architectural variants are largely analogous and can be transferred straightforwardly.

Let the model receive \(M\) relevant tokens in the clean setting, including the question text and evidence images, and \(M+N\) tokens when \(N\) irrelevant visual tokens are introduced. For notational convenience, we index the first \(M\) tokens as relevant and the remaining \(N\) tokens as irrelevant. We consider the attention layer within the decoder block, which serves as the primary module for token interactions across different positions. Such redundancy will disperse attention from relevant to irrelevant tokens and dilute effective information. Let \(h_x\) denote the hidden-state representation of the \(x\)-th token in the clean setting. When \(N\) irrelevant visual tokens are introduced, the corresponding hidden state is denoted by \(h'_x\):

\vspace{-15pt}
{\small
\begin{align}
    h_x =& \hat{h}_{x} + \alpha_1V_1 + \alpha_2V_2+\ldots+\alpha_MV_M = \hat{h}_{x} + \sum^{M}_{i = 1}\alpha_iV_i,\\
    h'_x =& \hat{h}'_{x}+\beta_1V'_1 + \beta_2V'_2+\ldots+\beta_{M+N}V'_{M+N}=\hat{h}'_{x}+\sum_{j=1}^{M+N}\beta_jV'_j,
\end{align}}
\vspace{-10pt}

\noindent where \(\hat{h}_{x}\) and \(\hat{h}'_{x}\) denote the outputs from the previous layer (due to the presence of residual connections), \(\alpha_i\) and \(\beta_i\) denote the attention weights, and \(V_i,V'_i \in \mathbb{R}^d\) are the \emph{value} vectors at the \(i\)-th position.

After redundant tokens are introduced, the attention mass originally allocated to the \(M\) relevant tokens is diluted and redistributed across the \(N\) irrelevant tokens. The average degree of attention dispersion can be expressed as
\begin{equation}
    \beta_i = \frac{\delta M}{M+N}\alpha_i,\quad (i = 1,2,\ldots,M),
\end{equation}
where \(\delta\) denotes the harmonization coefficient used in our approximate algorithm. Here, the dispersion ratio is positively correlated with \(\frac{M}{M+N}\). However, attention generally remains concentrated on the relevant tokens, so the actual degree of dispersion is typically smaller than \(\frac{M}{M+N}\), implying that \(\delta > 1\). Therefore, Eq.~2 can be rewritten as follows:

\vspace{-8pt}
{\small \begin{equation}
    h'_x=\hat{h}'_{x}+ \frac{\delta M}{M+N}\sum^{M}_{i = 1}\alpha_iV_i+\sum^{M+N}_{j=M+1}\beta_jV'_j.
\end{equation}}
% \vspace{-8pt}

We next compute the \textbf{Information Purity} $\eta$ of the hidden state of the token at the $x$-th position, defined as the proportion of its values attributable to relevant content, in order to quantify the contamination introduced by redundant tokens. The purity is formulated as:

\vspace{-8pt}
{\footnotesize\begin{equation}\begin{split}
    \eta &= \frac{h'_x-\sum^{M+N}_{j=M+1}\beta_jV'_j}{h'_x} = \frac{\hat{h}'_{x} + \frac{\delta M}{M+N}\sum^{M}_{i = 1}\alpha_iV'_i}{\hat{h}'_{x}+\frac{\delta M}{M+N}\sum^{M}_{i = 1}\alpha_iV'_i+\sum^{M+N}_{j=M+1}\beta_jV'_j} \\
    &=\frac{1+\frac{\hat{h}'_{x}}{\frac{\delta M}{M+N}\sum^{M}_{i = 1}\alpha_iV'_i}}{1+\frac{\hat{h}'_{x}}{\frac{\delta M}{M+N}\sum^{M}_{i = 1}\alpha_iV'_i}+\frac{M+N}{\delta M}\cdot \frac{\sum^{M+N}_{j=M+1}\beta_jV'_j}{\sum^{M}_{i=1}\alpha_iV'_i}}.\\
\end{split}
\label{eq:purity-main}
\end{equation}}
% \vspace{-8pt}

Here, \(\frac{\sum_{j=M+1}^{M+N}\beta_j V'_j}{\sum_{i=1}^{M}\alpha_i V'_i}\) denotes the ratio between the weighted value sum of the \(N\) redundant tokens and that of the \(M\) relevant tokens. In the extreme case, this ratio approaches \(\frac{N}{M}\). However, since relevant tokens typically receive larger attention weights than irrelevant ones, it cannot fully attain this upper bound. We therefore introduce a harmonization coefficient \(\lambda \in (0,1)\) to capture this bias, yielding \(\frac{\lambda N}{M}\). Accordingly, Eq.~5 can be simplified as:

\vspace{-8pt}
{\small \begin{equation} \begin{split}
    \eta =\frac{1+\frac{\hat{h}_x}{\sum^M_{i=1}\beta_iV'_i}}{1+\frac{\hat{h}_x}{\sum^M_{i=1}\beta_iV'_i}+\frac{M+N}{\delta M}\cdot \frac{\lambda N}{M}}\in (0,1),\quad \delta >1, \lambda \in (0,1).
\end{split} \end{equation}}
\vspace{-5pt}

At this point, we have derived the attenuation factor of information purity. However, due to space limitations, the discussion above presents only the central intuition of the analysis. For interested readers, the complete derivations beyond this point are provided in Appendix~\ref{theo}. Accordingly, we omit the technical details in the main text and state the final conclusion directly:

Information purity, denoted by \(\eta\), is recursively compositional. Specifically, after each block, the information purity is scaled by a factor of \(\eta\) relative to that of the preceding layer. Consequently, for an \(L\)-layer decoder, the information purity of the output tokens decreases to \(\eta^{L}\), implying an exponential compression of informative content. Moreover, as further verified in Appendix~\ref{theo}, a larger ratio \(\frac{N}{M}\) results in lower information purity.

This layer-wise proliferative contamination implies that, even when \(\eta\) is very close to \(1\), the resulting exponential decay can still cause substantial degradation. Such contamination weakens the model's ability to capture informative signals and may ultimately impair vocabulary decoding and text generation. The experiments presented in this paper further demonstrate the significant impact of this phenomenon on model performance.

Given the inevitability of redundancy in real-world data, we argue that greater emphasis should be placed on an LVLM's ability to sustain reasoning performance under redundant conditions, rather than under unrealistically clean settings, as an extended evaluation criterion. Our benchmark is motivated precisely by this perspective.

\subsection{Data Collection and Annotation}
\label{sec:question_type}

\begin{figure}
    \centering
    \includegraphics[width=0.7\linewidth]{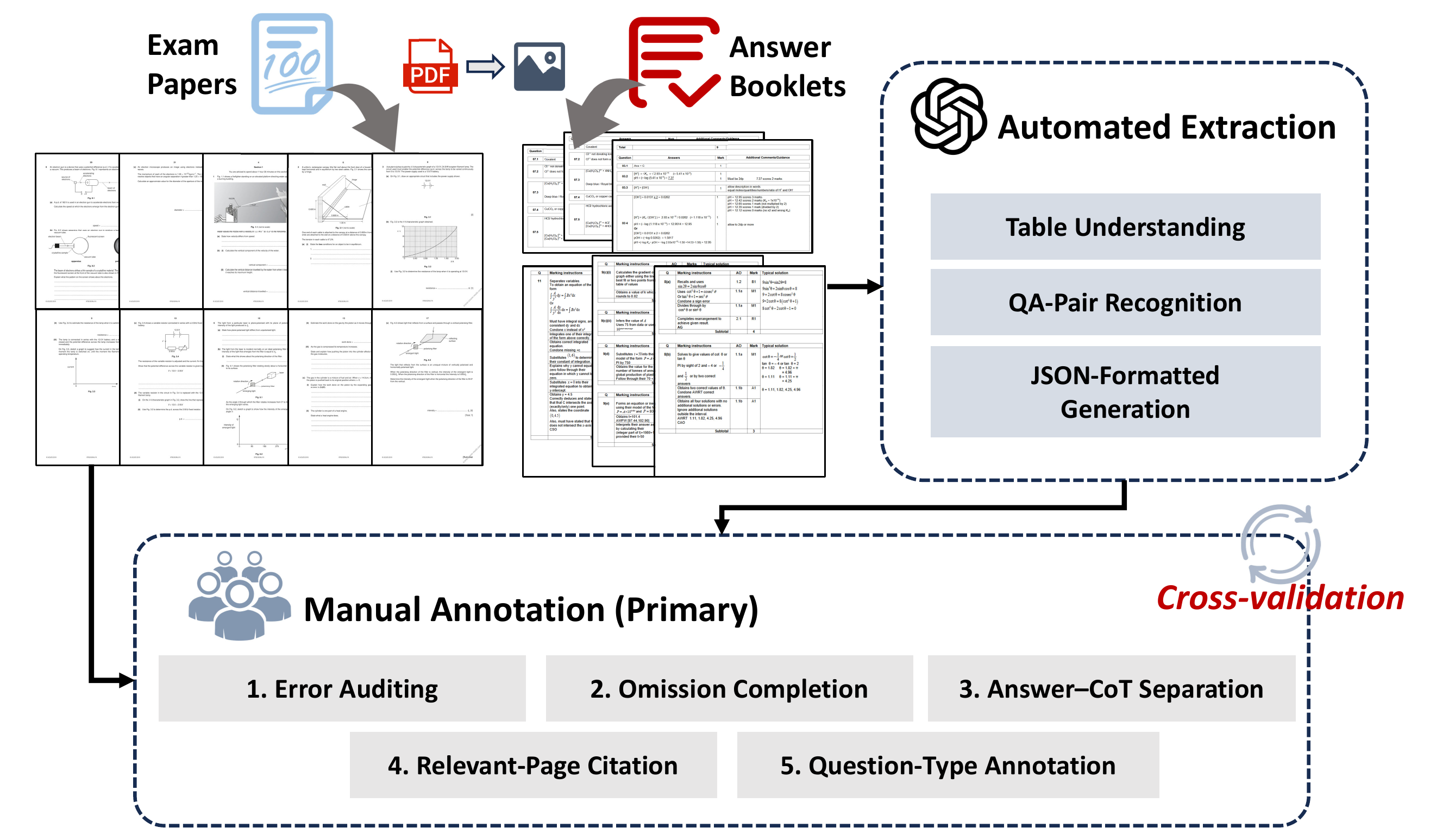}
    \caption{Data annotation process. An approach combining automation and manual efforts is used to balance efficiency and reliability.}
    \label{fig:anno}
\end{figure}

We seek a data source that combines reasoning complexity with large-scale document context, for which multi-page examination papers are a natural choice. Such documents feature diverse textual layouts and planar visual elements, requiring models to possess strong retrieval capabilities to identify relevant evidence, while the questions themselves demand nontrivial, human-level reasoning.

Focusing on three reasoning-intensive disciplines---mathematics, physics, and chemistry---we collected 995 exam PDFs from a publicly accessible website\footnote{\url{https://www.exam-mate.com}}. After excluding exams with limited visual content or only elementary-level difficulty, we retained 382 exams and their answer booklets. We then adopted a hybrid annotation strategy that combines manual annotation with LLM-assisted annotation to balance efficiency and quality control, as illustrated in Figure~\ref{fig:anno}.

Specifically, we first convert each PDF file into images to preserve complete visual information. We then use GPT-5 \cite{gpt5} to automatically remove blank or irrelevant pages, retaining only the core examination content and answer tables. Next, each answer-table page is processed by GPT-5 with layout-specific restrictive prompts, which instruct it to ignore distracting elements and generate a JSON object containing question indices and their corresponding answer text. We then manually complete missing cross-page question indices, decompose multi-part answers into structured lists, separate solution procedures from final answers, restore omitted text, and normalize all annotations into a standardized LaTeX format. At this stage, GPT-5 is used under strict constraints solely for text extraction as a format converter, thereby minimizing the influence of LLM preferences on dataset quality.

We categorized exam questions into six types: \textbf{Normal}, with fixed unique answers but flexible formatting; \textbf{Statement}, requiring free-form descriptive responses; \textbf{Calculate}, involving numerical computation with tolerance-based evaluation; \textbf{Choice}, selecting from predefined options; \textbf{Proof}, requiring step-by-step justification; and \textbf{Binary}, involving justified binary classification. Because automatic classification with GPT-5 yielded unsatisfactory results, we manually completed the annotation.

For each question, we annotate its relevant page range as a noise-free context (the ``Oracle'' setting). A page is considered relevant if it falls within the span from the beginning of the parent question stem to the end of the page on which the subquestion is fully presented. The total span is constrained to at most four pages. 

Ultimately, we construct a large-scale dataset comprising \textbf{10,242 QA pairs} and \textbf{7,155 high-resolution exam images}.

\subsection{Overview of Benchmark Key Highlights}
\begin{figure}
    \centering
    \includegraphics[width=0.5\linewidth]{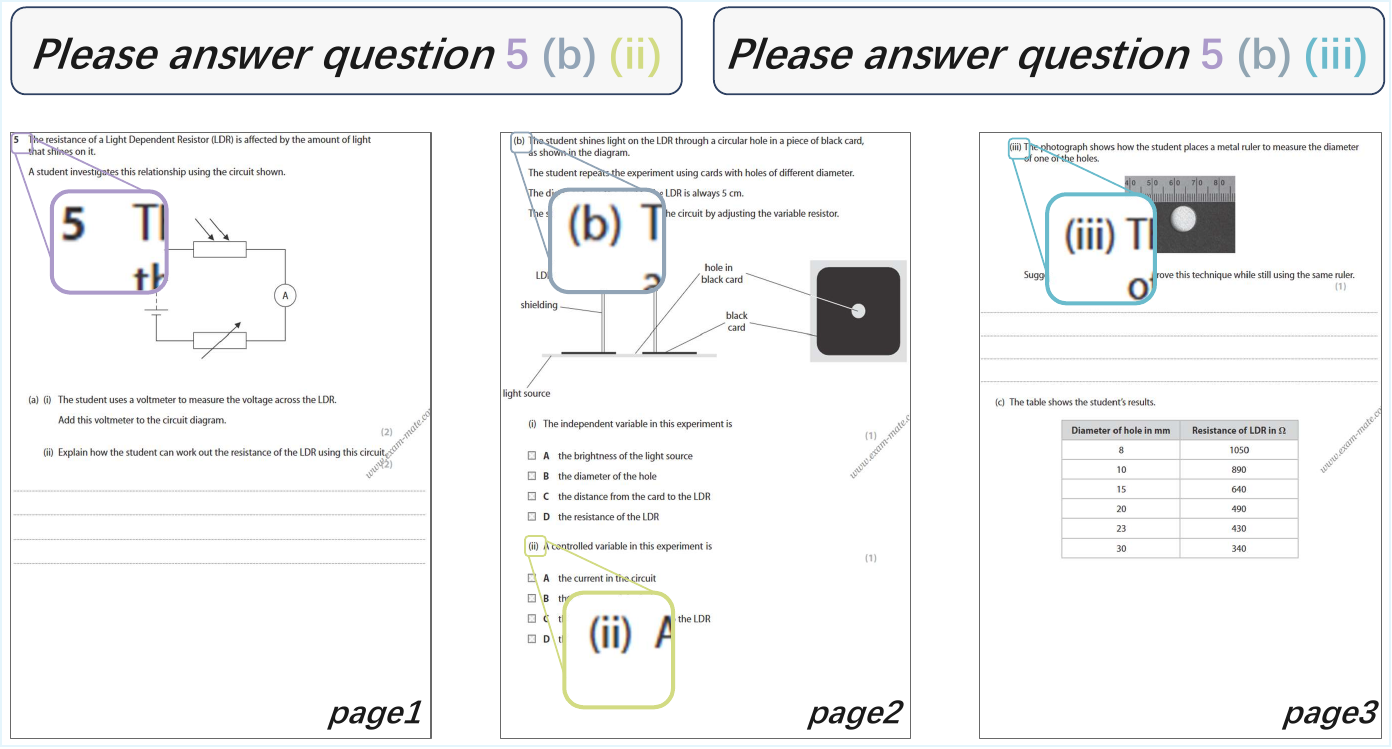}
    \caption{Example of nested question numbering in TestHallVQA, where subquestions are indicated by indentation and may span multiple pages.}
    \label{fig:cross}
\end{figure}

\begin{table}[ht]
    \centering
    \resizebox{\textwidth}{!}{%
    \begin{tabular}{l|cc|c|c|c|c|c}
    \toprule
       \multirow{2}{*}{\textbf{Datasets}}  & \multicolumn{2}{c|}{\textbf{Data Size}} & \multirow{2}{*}{\textbf{\ Question Types}} & \multirow{2}{*}{\textbf{Multi-Page}} & \multirow{2}{*}{\textbf{Deep-Reasoning}} & \multirow{2}{*}{\textbf{Evaluator}} & \multirow{2}{*}{\textbf{PVC}}\\
         & \ QA Pairs & \ Images &  &  &  &  & \\
         \midrule
         \multicolumn{8}{l}{ \color{burgundy} \textit{\textbf{DocVQA Tasks}}}\\
         SingleDocVQA \cite{docvqa}& 12,767& 50,000& 1(short) &{\color{red!90!black} \XSolidBrush} &{\color{red!90!black} \XSolidBrush} &ANLS \cite{anls}& {\color{red!90!black} \XSolidBrush}\\
         VisualMRC \cite{visualmrc}&10,197 &  30,562& 1(short) & {\color{red!90!black} \XSolidBrush}&{\color{red!90!black} \XSolidBrush} & BERTScore \cite{bertscore} &{\color{red!90!black} \XSolidBrush}\\
         ChartQA \cite{chartqa}&32,719 & 21,948&1(short) & {\color{red!90!black} \XSolidBrush}&{\color{red!90!black} \XSolidBrush} & EM (Exact Match) & {\color{red!90!black} \XSolidBrush}\\
         MP-DocVQA \cite{MP}& 46,000 & 48,000 & 1(short) &  {\color{green!60!black} \Checkmark} & {\color{red!90!black} \XSolidBrush} & ANLS & {\color{red!90!black} \XSolidBrush}\\
         DUDE \cite{dude}& 28,000 & 41,541 & 1(short) & {\color{green!60!black} \Checkmark} & {\color{red!90!black} \XSolidBrush} & ANLS & {\color{red!90!black} \XSolidBrush}\\
         MMLongBench-Doc \cite{mmlongbench}&1,082 & 6,412& 1(short) & {\color{green!60!black} \Checkmark} &{\color{red!90!black} \XSolidBrush} & regex-based matching & {\color{red!90!black} \XSolidBrush}\\
         LongDocURL \cite{longdocurl}&2,325 & 33,898& 1(short) &{\color{green!60!black} \Checkmark} & {\color{red!90!black} \XSolidBrush}& regex-based matching &{\color{red!90!black} \XSolidBrush}\\
         \multicolumn{8}{l}{ \color{burgundy} \textit{\textbf{Scientific VQA Tasks}}}\\
         MathVista \cite{mathvista}&6,141 &  5,487 &2(short\textsubscript{45\%}, choice\textsubscript{55\%}) & {\color{red!90!black} \XSolidBrush}&{\color{green!60!black} \Checkmark} & regex-based matching&{\color{red!90!black} \XSolidBrush}\\
         MathVerse \cite{mathverse}& 2,612 & 2,420 & 2(short\textsubscript{38\%}, choice\textsubscript{62\%}) & {\color{red!90!black} \XSolidBrush}& {\color{green!60!black} \Checkmark}&GPT-4V \cite{gpt4v_systemcard} &{\color{red!90!black} \XSolidBrush}\\
         MMMU \cite{mmmu}& 11,550& 11,264& 2(short\textsubscript{6\%}, choice\textsubscript{94\%}) & {\color{red!90!black} \XSolidBrush}& {\color{green!60!black} \Checkmark}& regex-based matching & {\color{red!90!black} \XSolidBrush}\\
         MMMU-Pro \cite{mmmu-pro}&  3,460& -& 1(choice)& {\color{red!90!black} \XSolidBrush}&{\color{green!60!black} \Checkmark} & regex-based matching&{\color{red!90!black} \XSolidBrush }\textsubscript{50\%} {\color{green!60!black} \Checkmark}\textsubscript{50\%}\\
         SceMQA \cite{scemqa}&1,045 & -&2(short\textsubscript{20\%}, choice\textsubscript{80\%}) & {\color{red!90!black} \XSolidBrush}&{\color{green!60!black} \Checkmark} &EM (Exact Match)  &{\color{red!90!black} \XSolidBrush}\\
         HC-M3D \cite{HC-M3D}& 1,851&1,851 &1(choice) &{\color{red!90!black} \XSolidBrush} &{\color{green!60!black} \Checkmark} & - & {\color{red!90!black} \XSolidBrush}\\
         \midrule
         TestHallVQA (\textit{ours})&10,242 & 7,155& 6 (Details in Caption) & {\color{green!60!black} \Checkmark}& {\color{green!60!black} \Checkmark}&LLMs (open-source friendly) &{\color{green!60!black} \Checkmark}\\
         \bottomrule
    \end{tabular}%
    }
    \caption{Comparison of TestHallVQA with existing related benchmarks. ``PVC'' denotes Pure Visual Cues. For question types, ``short'' refers to short-answer questions, and ``choice'' refers to multiple-choice questions. The six question types in TestHallVQA are: Normal, Statement, Choice, Calculate, Proof, and Binary.}
    \label{tab:comp}
\end{table}

\textbf{Cross-Page Challenges.} TestHallVQA exam papers adopt a hierarchical question--subquestion structure, in which a substantial portion further employs nested numbering and indentation, distributing complete question identifiers across multiple pages (as shown in Figure~\ref{fig:cross}). This structure renders page-level RAG methods \cite{Rag1,Rag3} ineffective and requires LVLMs to infer inter-page dependencies and integrate dispersed evidence through global reasoning.

\textbf{Pure Visual Cues.}
Following \cite{mmmu-pro}, humans acquire both textual and graphical information through vision, suggesting that LVLMs should likewise rely solely on \textbf{``pure visual cues''} rather than explicit textual input. Accordingly, TestHallVQA adopts a question-index--guided prompting strategy (e.g., ``Please answer question 3(b)ii.''), where no semantically relevant text is provided. LVLMs must therefore gather all evidence through visual perception alone. This end-to-end VQA setting evaluates their intrinsic visual analysis and text reasoning abilities while avoiding confounding factors from system-level pipelines.

\textbf{Open-source-friendly LLMs as judges.} Most existing planar VQA benchmarks rely on heuristic evaluation metrics (e.g., rule-based matching and ANLS \cite{anls}), which are brittle to semantically equivalent variations that preserve the target format. This limitation may systematically underestimate model performance and, in turn, overstate the difficulty of the benchmark. Following prior work \cite{judge2,judge3,judge4}, we adopt an LLM-as-a-Judge evaluation framework with format-specific judging protocols tailored to the six task types in TestHallVQA. Subsequent experiments show that this setup enables LLMs, including open-source models, to achieve strong agreement with human preferences. Complete evaluation templates and illustrative examples are provided in Appendix~\ref{sec:judge}.

We compare TestHallVQA with related benchmarks in Table~\ref{tab:comp}.

\subsection{Multi-Level Redundant Context Injection}
The objective of the TestHallVQA benchmark is to unify the evaluation of problem-solving capacity (Reasoning) and robustness to redundant visual contamination (Retrieval) into a coupled capability.

Reasoning capability can be evaluated directly through answer accuracy. In contrast, assessing retrieval capability requires more careful design. One intuitive approach is to ask the model to identify the relevant pages and use page prediction accuracy as the retrieval metric \cite{dude,MP}. However, this formulation effectively reduces the benchmark to a decoupled retrieval-and-reasoning setting, where the two capabilities are evaluated independently. Instead, retrieval capability should be measured by performance under increasing visual redundancy: models that remain accurate in highly redundant contexts exhibit stronger retrieval capability.

To introduce such progressively increasing visual redundancy, we propose a controllable \textbf{redundant context injection} strategy. Specifically, each question is paired with multiple context scales, termed \textbf{``regimes''}. The first regime includes only the relevant pages (i.e., the ``Oracle'' setting), and each subsequent regime adds ten redundant pages until the full exam document is reached. To maximize confusability, redundant pages are sampled from the same exam paper. Given that the maximum page count in our collected exams is 44, the number of regimes is capped at five. Detailed pseudocode is provided in Appendix~\ref{sec:regimes}.

\begin{figure}
    \centering
    \includegraphics[width=0.5\linewidth]{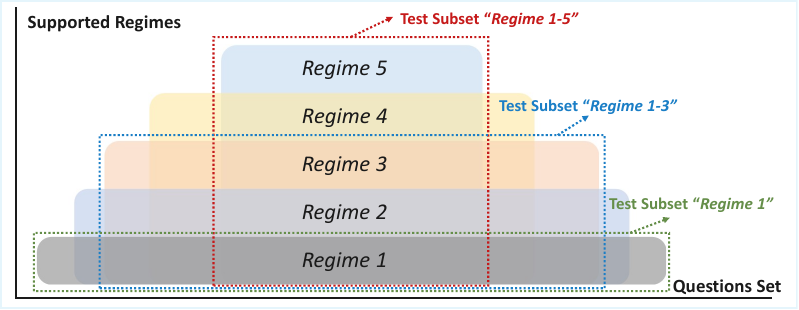}
    \caption{Pyramid-shaped distribution in TestHallVQA. The boxed region denotes a subset in which the question set remains identical across all regimes within the subset.}
    \label{fig:testset}
\end{figure}

However, due to the varying lengths of the documents, TestHallVQA's overall distribution follows a stepped pattern, as shown in Figure~\ref{fig:testset}. We partition TestHallVQA into five test subsets using the rectangular selection method illustrated in Figure~\ref{fig:testset}. This design maximizes data utilization while ensuring that the questions in each test subset remain consistent across regime transitions.

\subsection{\texorpdfstring{F1-R\textsuperscript{2}}{F1-R2} Metric}
\label{sec:F1R2}

After constructing the multi-regime evaluation subsets described above, we now detail the computation of our \textbf{F1-R\textsuperscript{2}} metric (\textbf{F1} score for \textbf{R}etrieval and \textbf{R}easoning). For clarity, we take the computation of F1-R\textsuperscript{2} on the subset ``Regime 1--3'' as an example; the other cases are defined analogously.

Each question in the subset ``Regime~1--3'' is evaluated under different regimes using the LLM-as-a-Judge protocol, in which the LLM assigns a correctness score in the range of $[0,1]$ to each model output. The average accuracy scores across the three regimes are denoted by \(e_1, e_2, e_3 \in [0,100]\), expressed as percentages. Let \(\bar{p}_1, \bar{p}_2, \bar{p}_3\) denote the average numbers of pages under the corresponding regimes. The \mbox{F1-R\textsuperscript{2}} metric is defined as follows:

\vspace{-5pt}
{\small
\begin{equation}
    \mathrm{F1\text{-}R}^2 = \frac{2 \cdot S_{RT} \cdot S_{RN}}{S_{RT} + S_{RN}},
\end{equation}
}

\noindent where \(S_{RT}\) denotes the retrieval score, and \(S_{RN}\) denotes the reasoning score. \(S_{RN} \in [0,1]\) denotes the accuracy under the last regime (Regime 3 in this case), reflecting the model's true reasoning ability in document-level, realistically redundant settings:

\vspace{-7pt}
{\small
\begin{equation}
    S_{RN} = e_3 \% = \frac{e_3}{100} \in [0,1].
\end{equation}
}
\vspace{-10pt}

\begin{figure}
    \centering
    \includegraphics[width=0.5\linewidth]{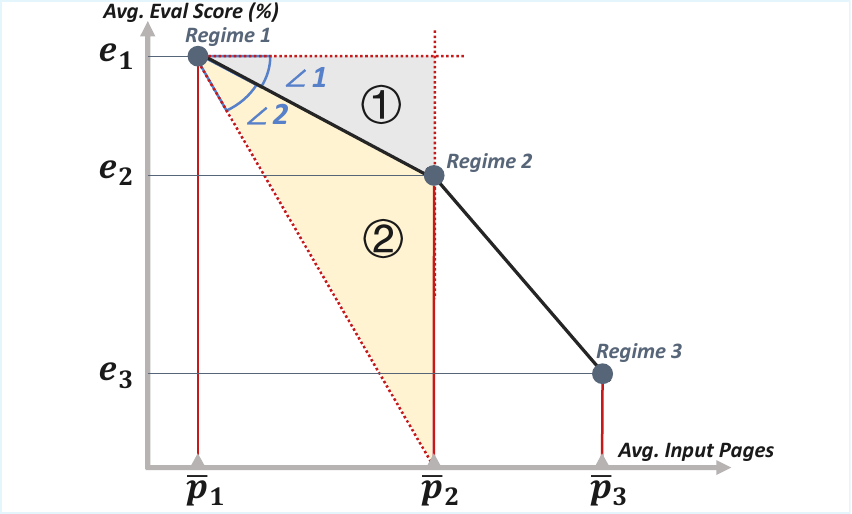}
    \caption{Performance decay of an LVLM across the three contextual regimes in TestHallVQA.}
    \label{fig:line}
\end{figure}

For \(S_{RT}\), we define it as the slope of performance degradation induced by the introduction of redundant pages. To ensure that \(S_{RT}\) is sensitive to performance degradation and possesses sufficient discriminative power for model comparison, we derive it from the \textbf{rotation angle}, which serves as a measure of the extent of decay. Subsequent ablation studies further validate its superior discriminative capacity.

Specifically, as shown in Figure~\ref{fig:line}, when the first segment of the decay polyline reaches its maximum drop \((e_2 = 0)\), it removes an area represented by the triangles \(\circled{1} + \circled{2}\). At this point, the rotation angle of the score polyline is the sum of \(\angle 1 + \angle 2\). In contrast, when retrieval performance is perfect, the polyline remains flat, and no area is removed. Using this bounded interval, the normalized retrieval performance is defined as:

\vspace{-5pt}

{\small
\begin{equation}
    k_1 = \frac{\angle{2}}{\angle{1} + \angle{2}} =1 - \frac{\angle{1}}{\angle{1} + \angle{2}}
= 1 -\frac{ \arctan (\frac{e_1-e_2}{\bar{p}_2 - \bar{p}_1}) }{\arctan (\frac{e_1}{\bar{p}_2 - \bar{p}_1})} \in [0,1].
\end{equation}
}
\vspace{-5pt}

Similarly, the decay factor from Regime 2 to 3 is:

{\small
\begin{equation}
    k_2 = 1 -\frac{ \arctan (\frac{e_2-e_3}{\bar{p}_3 - \bar{p}_2}) }{\arctan (\frac{e_2}{\bar{p}_3 - \bar{p}_2})} \in [0,1].
\end{equation}
}
\vspace{-5pt}

The retrieval score \(S_{RT}\) is then computed as a weighted combination of \(k_1, k_2\), where the weights correspond to the average page span between the two regimes:

\vspace{-5pt}
{\small
\begin{equation}
    S_{RT} = \frac{(\bar{p}_2 - \bar{p}_1) k_1 + (\bar{p}_3 - \bar{p}_2) k_2}
    {\bar{p}_3 - \bar{p}_1} \in [0,1].
\end{equation}
}

\section{Experiments}
\subsection{Evaluation Protocols}

We evaluate a broad range of LVLMs officially released by January 2026 on each test subset of TestHallVQA. We also include OCR-based baselines as supplementary references. Specifically, \texttt{PaddleOCR-v5} \cite{paddleocr} is used to extract text from the original images and organize it in natural reading order. The extracted text is then provided as contextual input in place of the images.

We use \texttt{Qwen3-30B-A3B-Instruct-2507} \cite{qwen3} as the judge LLM, employing the standard prompt templates described in Appendix~\ref{sec:judge}. We also gratefully utilize the vLLM framework \cite{vllm} to enhance inference efficiency. We adopt greedy decoding for reproducibility of results and perform all experiments on 8 A100-SXM4-80GB GPUs.

Given the large evaluation scale and broad model coverage, we evaluate API-based models and models with more than 70B parameters on a \emph{mini} test set that matches the full test-set distribution (see Appendix~\ref{sec:stati}), balancing fidelity and computational cost.

Some models are unable to accommodate the context length introduced by multi-page image inputs under certain evaluation settings. Their results on the corresponding subsets are therefore marked as \textbf{null}, indicating insufficient long-context capability. Nevertheless, to improve the completeness of the experimental evaluation, we additionally report results at lower image resolutions to fill in these \textbf{null} entries wherever reasonable. Specifically, when a subset exceeds a model's context budget, we uniformly downsample all images in that subset until all samples become processable, using scaling factors of $0.707$ (50\% pixel reduction) and $0.5$ (75\% pixel reduction). If neither setting suffices, the subset is excluded. Importantly, \textbf{only results obtained at the same image resolution are horizontally comparable}, since varying resolutions introduce confounding effects from differences in visual fidelity.

\subsection{Main Results}

\begin{table}[ht]
    \centering
    \resizebox{\textwidth}{!}{
    \renewcommand{\arraystretch}{1.1}
    \begin{tabular}{l|c|c|ccc|ccc|ccc|ccc}
    \toprule
         \multirow{2}{*}{\textbf{Model}} &\multirow{2}{*}{\textbf{Size}} & \textbf{Oracle} & \multicolumn{3}{c}{\textbf{Regime 1--2 (\(p_{\max} = 10\))}} & \multicolumn{3}{|c}{\textbf{Regime 1--3 (\(p_{\max} = 20\))}} & \multicolumn{3}{|c}{\textbf{Regime 1--4 (\(p_{\max} = 30\))}} & \multicolumn{3}{|c}{\textbf{Regime 1--5 (\(p_{\max} = 40\))}}\\
         \cmidrule(){3-15}
         &&  $S_{RN}$ &  $S_{RT}$ & $S_{RN}$ & $F1-R^2$ & $S_{RT}$ & $S_{RN}$ & $F1-R^2$ & $S_{RT}$ & $S_{RN}$ & $F1-R^2$ & $S_{RT}$ & $S_{RN}$ & $F1-R^2$ \\
         \midrule
        \rowcolor{gray!15} \multicolumn{15}{l}{ \color{burgundy} \textit{\textbf{LLMs/LVLMs with OCR-processed text input}}} \\
         \rowcolor{gray!15}
         GPT-5 \citeyear{gpt5}&-&46.68& 68.85&42.62 & 52.65 &74.37 & 41.34 & 53.14 & 70.62& 39.74& 50.86& 71.42 &38.40 &49.95\\
         \rowcolor{gray!15}
         Qwen2.5-VL \citeyear{qwen2.5vl}&7B& 28.74& 69.14 &24.96 & 36.68 & 72.15 & 23.07 &  34.96&   64.87 & 20.49&31.14 &  55.41&17.92 &27.08\\
         \rowcolor{gray!15}
         InternVL3.5 \citeyear{internvl3d5}&8B& 39.08&  78.02& 36.24& 49.49 & 70.19 & 32.16 &44.11  &  67.55 & 27.42& 39.07&  69.45&25.18 &36.96\\
         \rowcolor{gray!15}
         Mistral-Small-3.1 \citeyear{mistral}&24B& 40.24 & 80.62 & 38.85& 52.43& 70.28& 34.33& 46.13&71.34 & 30.47& 42.65& 72.86& 28.04&40.50\\
         \rowcolor{gray!15}
         DeepSeek-R1 \citeyear{deepseekr1}&32B& 42.85 &  80.11& 40.41&53.72 & 78.44 &37.04 &50.32 &64.57 & 33.64&44.23 &70.83 & 28.44& 40.58\\

         \multicolumn{15}{l}{ \color{burgundy} \textit{\textbf{Closed-Source LVLMs}}} \\
         Qwen-VL-max$^{\ast}$ \citeyear{qwenvl}&-&65.76& 70.29&61.62 & 65.67 &62.30 & 55.34 &58.61 &61.90 & 51.46&56.19&58.32 &44.70 &50.60\\
         GPT-4o$^{\ast}$ \citeyear{gpt40}&-&78.91&59.45 &72.87 &65.47& 60.05 &67.30&63.46&58.49  &62.38 &60.37 &58.83 &55.34&54.01\\
         GPT-5$^{\ast}$ \citeyear{gpt5}&-& \underline{\textbf{85.77}} & 63.57 &80.43 &71.01& 67.96 &\underline{\textbf{76.59}} &\underline{\textbf{72.01}}& 63.25& 70.81& 66.81& \textbf{61.45}& \underline{\textbf{65.59}}&\underline{\textbf{63.45}}\\
         Gemini-2.5-Pro$^{\ast}$ \citeyear{gemini}&-&84.50& \textbf{73.95}&\underline{\textbf{80.82}} &\underline{\textbf{77.23}} & \textbf{68.01}&75.59 &71.59 &\underline{\textbf{67.93}} &\underline{\textbf{71.66}} &\underline{\textbf{69.74}}& 61.64& 64.36&62.97\\

         \multicolumn{15}{l}{ \color{burgundy} \textit{\textbf{Large-Scale Open-Source LVLMs}}} \\
         LLaVA-OneVision$^{\ast}$ \citeyear{llavaov}&72B& 32.44 &43.36$\ddagger$&15.69$\ddagger$&23.04$\ddagger$&-&-&-&-&-&-&-&-&-\\
         LLaVA-Next$^{\ast}$ \citeyear{llavanext}&72B&29.29 &74.60 & 26.14&38.71 & 67.80$\dagger$&22.67$\dagger$ &33.97$\dagger$ &67.13$\ddagger$ & 14.98$\ddagger$&24.49$\ddagger$ &53.66$\ddagger$ & 8.84$\ddagger$ &15.17 $\ddagger$\\
         InternVL3$^{\ast}$ \citeyear{internvl3}&78B&  64.54 & 65.84 & 59.77 &62.65 &60.66$\dagger$& 53.64$\dagger$&56.93 $\dagger$&55.59 $\dagger$ &  47.78 $\dagger$ &51.38$\dagger$ & 54.58 $\ddagger$ &37.24 $\ddagger$ &44.27 $\ddagger$ \\
         Qwen2.5-VL$^{\ast}$ \citeyear{qwen2.5vl}&72B& 63.20 & 54.84 & 56.42 & 55.61& 50.11 & 48.68 &49.38 & 47.02 & 42.43 &44.68 & 52.45 & 36.81 &43.26 \\
         Mistral-Small-3.1 \citeyear{mistral}&24B& 65.20 & 74.15& 61.63& 67.31&64.24 &54.82 &59.15 &\textbf{65.11} & 49.55& 56.27&57.25 & 43.92&49.70\\
         InternVL3.5 \citeyear{internvl3d5}&38B& 72.30&\textbf{77.82}&69.23&\textbf{73.21}&\textbf{71.33}&65.06&\textbf{68.05}&68.72$\dagger$&57.47$\dagger$&62.59$\dagger$&60.77$\dagger$&51.85$\dagger$&55.96$\dagger$\\
         Qwen3-VL \citeyear{Qwen3-VL}&30B& \textbf{80.65} &  58.22 & \textbf{74.42} & 65.33 & 64.13 & \textbf{70.36} & 67.10& 61.79 & \textbf{65.53} & \textbf{63.61}& \textbf{62.56} & \textbf{60.12} &\textbf{61.37}  \\

         \multicolumn{15}{l}{ \color{burgundy} \textit{\textbf{Small-Scale Open-Source LVLMs}}} \\
         InternLM-XC2.5 \citeyear{ixc}&7B& 25.75 & 42.14 & 17.90& 25.12 & 29.25$\dagger$&10.52$\dagger$&15.47$\dagger$ & 37.76$\ddagger$&6.61$\ddagger$ &11.25$\ddagger$ & 40.50$\ddagger$ & 3.20$\ddagger$ & 5.93$\ddagger$ \\
         LLaVA-OneVision \citeyear{llavaov}&7B &23.67 &52.77$\ddagger$& 13.12$\ddagger$ & 21.01$\ddagger$ & -&-&-&-&-&-&-&-&-\\
         LLaVA-Next \citeyear{llavanext}&8B&17.57 & 28.63$\ddagger$ & 8.36$\ddagger$ &12.94$\ddagger$&-&-&-&-&-&-&-&-&-\\
         DeepSeek-OCR \citeyear{deepseekocr}&3B& 18.49 &36.22$\dagger$ & 10.62$\dagger$& 16.42$\dagger$ & 16.97$\ddagger$ & 3.14$\ddagger$ & 5.29$\ddagger$& 14.50$\ddagger$ & 1.24$\ddagger$ & 2.28$\ddagger$ & -&-&-\\
         Qwen2.5-VL \citeyear{qwen2.5vl}&7B &56.04&38.41&45.73&41.75&43.67&38.68&41.02&47.49&31.80&38.09&54.21&24.66&33.90  \\
         InternVL3 \citeyear{internvl3}&8B & 60.41&70.55&56.38&62.67&68.33$\dagger$&47.50$\dagger$&56.04$\dagger$&60.82$\dagger$&39.77$\dagger$&48.09$\dagger$&62.28$\ddagger$&26.12$\ddagger$&36.80$\ddagger$\\
         GLM-4.1V-Thinking \citeyear{glm}&9B & \textbf{79.94} &63.21&\textbf{74.75}&68.49& 60.27&65.80&62.91& 67.31$\dagger$&54.18$\dagger$&60.03$\dagger$& 60.83$\dagger$&49.26$\dagger$&54.44$\dagger$\\
         InternVL3.5 \citeyear{internvl3d5}&8B& 69.43 & \underline{\textbf{78.60}} & 66.49& \textbf{72.04}& \underline{\textbf{74.94}} & 62.66 & \textbf{68.25}& 70.58$\dagger$ & 54.42$\dagger$ & 61.45 $\dagger$&66.29 $\dagger$ & 49.70 $\dagger$ & 56.81$\dagger$  \\
          Qwen3-VL \citeyear{Qwen3-VL}&8B&78.28&61.72&72.23&66.56&  66.71 &\textbf{68.86}& 67.77  & \textbf{67.41}&\textbf{62.88}& \textbf{65.07}&\underline{\textbf{67.16}}&\textbf{55.55}&\textbf{60.81} \\

    \bottomrule
    \end{tabular}}
    \caption{Main results on TestHallVQA. ``Oracle'' denotes Regime 1, which represents the clean case with relevant pages as context. ``Regime~1--N'' denotes the test subset comprising Regime~1 through Regime~N. ``\(p_{\max}\)'' represents the maximum number of images across the selected regimes. ``\textbf{$\ast$}'' indicates evaluation on the \emph{mini-test} set. ``\textbf{$\dagger$}'' and ``\textbf{$\ddagger$}'' indicate image side-length scaling by 0.707 (50\% pixels) and 0.5 (25\% pixels), respectively. Bold values indicate the best result among models of comparable scale, while bold and underlined values denote the global SOTA (excluding cases with OCR processing and resolution reduction).}
    
    \label{tab:main}
\end{table}

Table~\ref{tab:main} summarizes the evaluation results of all models across the five test subsets of TestHallVQA. Based on a detailed analysis of the experimental results, we report the following key observations:

\textbf{Reducing image resolution leads to performance degradation.}
For certain LVLMs, such as the LLaVA series, input-length constraints remain prohibitive even after the image resolution is reduced to one quarter of the original pixel count. Moreover, when comparing the resolution-reduced subset with the immediately preceding subset evaluated at the original resolution, we observe a sharp drop in both $S_{RT}$ and $S_{RN}$. Several failure cases indicate that this degradation stems from the fact that fine-grained visual elements, such as subscripts and other small typographic markers, become indistinguishable after resolution reduction.

\textbf{Pure-text inputs are insufficient.}
As shown in the gray rows of Table~\ref{tab:main}, OCR-based conversion discards graphical elements, charts, and other visual cues, leading to substantial evidence loss and a lower $S_{RN}$ score. This result underscores the strong visual dependence of TestHallVQA. Although converting the input into pure text reduces the token count and alleviates context-length pressure, thereby yielding a high $S_{RT}$, this advantage is of limited practical value given the extremely low answer accuracy.

\textbf{Reasoning performance deteriorates sharply under multi-page redundancy.}
Focusing on the \(S_{RN}\) metric, we observe that models perform strongly when conditioned solely on the ``Oracle'' regime, representing a highly idealized and noise-free setting (e.g., GPT-5 achieves 85.77\%). However, extending the context to the document level results in a sharp and monotonic decline in \(S_{RN}\) across all models (e.g., GPT-5 drops to 63.45\% on the ``Regime 1--5'' subset). This trend is fully consistent with the theoretical prediction in Section~\ref{sec:theo}, which states that information purity decreases as \(\frac{N}{M}\) increases. The pronounced left--right contrast in the table underscores that existing LVLMs experience substantial degradation in reasoning performance under complex contextual settings and demonstrate limited robustness to redundancy.

\textbf{Retrieval capacity saturates as the number of pages increases.}
As shown throughout the table, robustness to redundant context, as measured by $S_{RT}$, remains consistently low across all models and further deteriorates as the evaluation regimes scale up, revealing a fundamental retrieval bottleneck. These findings suggest that, for complex visual tasks, the effective context length of LLMs is substantially smaller than that implied by needle-in-a-haystack evaluations~\cite{needle}. Therefore, metrics derived from the severity of performance degradation, such as $S_{RT}$, are better suited to assessing context capacity.

\textbf{Counterintuitive finding: scaling does not mitigate contamination.}
Intuitively, larger models are expected to be more robust to redundant-token contamination and to better preserve the accuracy achieved on the oracle set. However, our results suggest that this expectation does not hold in practice: larger models often attain comparable or even lower $S_{RT}$ than their smaller counterparts (e.g., InternVL and Qwen across different parameter scales), and even advanced closed-source models rarely achieve the best overall $S_{RT}$. This phenomenon can be interpreted through the information-purity formulation $\eta^{L}$ introduced in Section~\ref{sec:theo}. Specifically, although increasing the number of layers may allow each layer to capture more fine-grained features and partially suppress contamination, deeper models with larger $L$ tend to yield lower $\eta^{L}$, thereby exacerbating the propagation of noise introduced by redundant tokens.

\subsection{Extended Analysis and Ablations}

\begin{table}[t]
\centering
\scriptsize
\setlength{\tabcolsep}{4pt}
\begin{tabularx}{\columnwidth}{c|XXXX}
\toprule
\diagbox[width=6em]{Evaluator}{Model}
& Gemini-2.5-Pro
& InternVL3.5-38B
& GLM-4.1V-9B 
& InternVL3.5-8B \\

\midrule
Human Avg.      
& 0.631
& 0.514 
& 0.593 
& 0.547\\

GPT-5      
& 0.634 (+0.003) 
& 0.520 (+0.006) 
& 0.596 (+0.003)
& 0.551 (+0.004)\\

GPT-5-nano 
& 0.629 (-0.002) 
& 0.524 (+0.01) 
& 0.595 (+0.002)
& 0.544 (-0.003)\\

Qwen3-30B  
& 0.638 (+0.007) 
& 0.517 (+0.003) 
& 0.585 (-0.008)
& 0.543 (-0.004)\\

Qwen3-8B   
& 0.609 (-0.022) 
& 0.499 (-0.015) 
& 0.587 (-0.006)
& 0.537 (-0.01)\\
\bottomrule
\end{tabularx}
\caption{LLM-as-a-Judge evaluation results compared with human annotations on 1K randomly sampled instances from different regimes of the mini-test set.}
\label{tab:judge_diff}
\end{table}

\begin{figure}[t]
    \centering
    \includegraphics[width=0.8\linewidth]{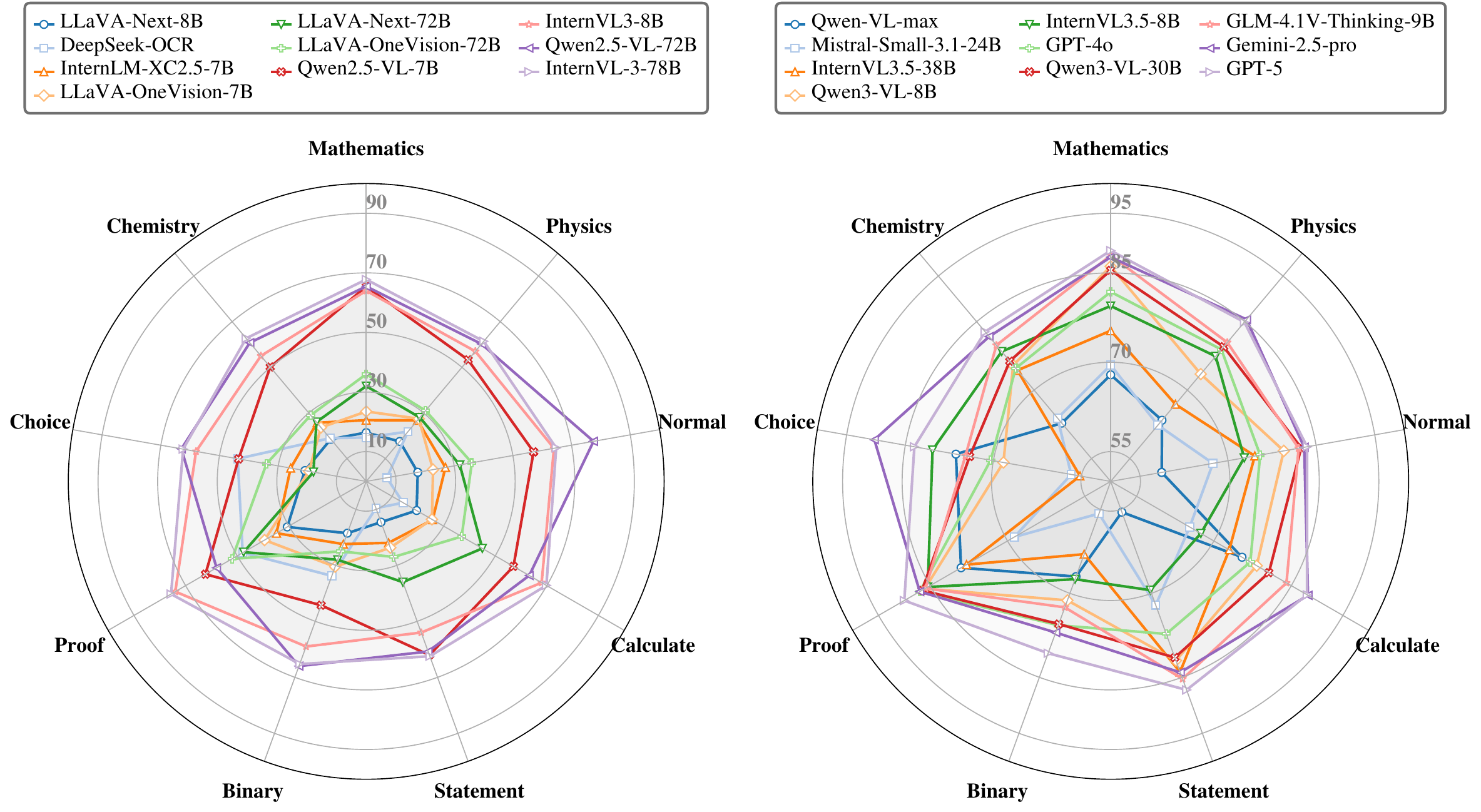}
    \caption{Classification accuracy radar plots, divided into two figures according to score ranking.}
    \label{fig:leida}
\end{figure}

First, we validate our LLM-as-a-Judge paradigm by comparing the accuracy scores assigned by human evaluators and LLM-based judges, including two human annotators and four representative LLMs. As shown in Table~\ref{tab:judge_diff}, aside from the weaker performance of the smaller Qwen3-8B, models at other scales demonstrate strong agreement with human judgments. This indicates that our prompt design (Appendix~\ref{sec:judge}) provides clear criteria for fine-grained LLM scoring while mitigating the influence of self-preference.

Figure~\ref{fig:leida} reports the accuracy of selected LVLMs across different question categories under Regime~1, the broadest test subset of the benchmark. Detailed statistics for each question category are provided in Appendix~\ref{sec:stati}. The results indicate that the difficulty levels across the various disciplines and question types are relatively well balanced within the benchmark.

\begin{figure}
    \centering
    \includegraphics[width=0.5\linewidth]{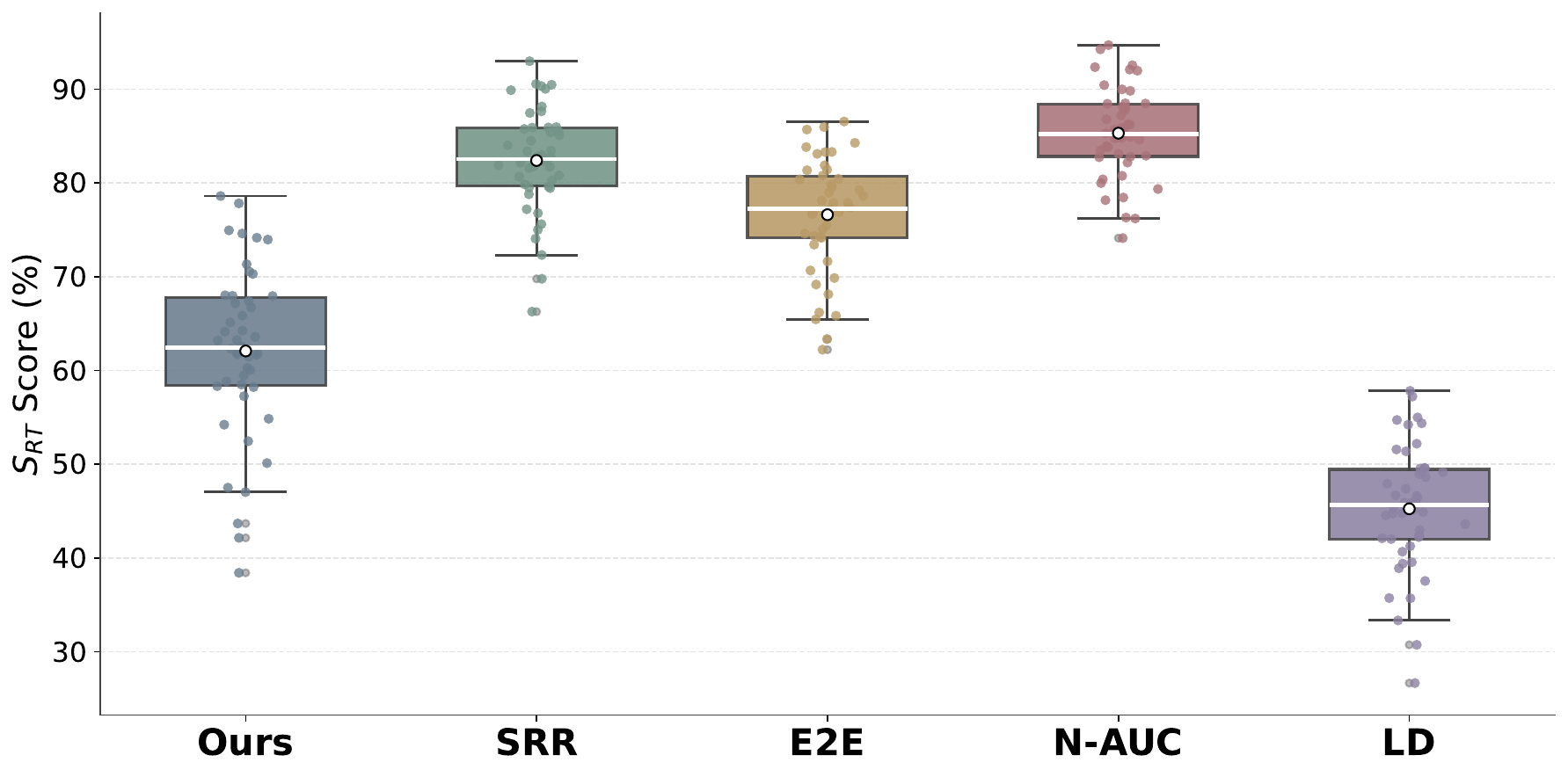}
    \caption{Box plots of the score distributions of $S_{RT}$ under the rotation-angle method and baseline methods.}
    \label{fig:boxplot}
\end{figure}

To demonstrate the advantages of the rotation angle-based $S_{RT}$ metric, we compare it against several baseline metrics, including step retention ratio (SRR), end-to-end retention (E2E), normalized AUC (N-AUC), and log-based decay (LD); their formal definitions are provided in Appendix~\ref{sec:SRT}. Figure~\ref{fig:boxplot} shows the distributions of $S_{RT}$ scores across all models and subsets under these metrics, excluding the OCR and image-resolution experiments. As illustrated, even after normalization, the baseline metrics compress models with substantially different capabilities into a narrow score range. In contrast, our metric demonstrates the highest sensitivity, producing the widest score distribution (from 0.38 to 0.79) and thereby offering the strongest discriminative power for the evaluation pipeline.

\begin{figure}[t]
    \centering
    \includegraphics[width=0.5\linewidth]{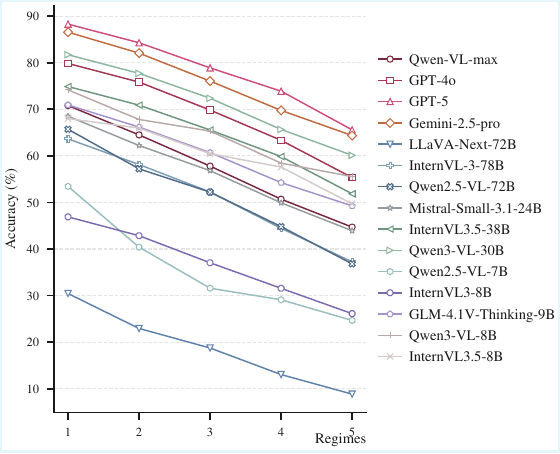}
    \caption{Performance degradation of selected LVLMs on the ``Regime 1--5'' Subset.}
    \label{fig:curve}
\end{figure}

Figure~\ref{fig:curve} shows the degradation curves of selected LVLMs under different regimes on the most comprehensive test subset, ``Regime~1--5.'' The steep declines provide an intuitive illustration of the models' substantial performance deterioration, highlighting the detrimental impact of redundant visual contamination.

We provide more detailed supplementary experiments and analyses for interested readers. Appendix~\ref{sec:extend} presents extended ablation studies on the effects of the number of pages involved in each question and the degree of visual confusability. In Appendix~\ref{sec:limit}, we further show that existing visual encoders are unable to capture the sequential temporal order across multiple images; this structural limitation may further aggravate redundant visual contamination.

\section{Error Analysis and Case Studies}
\begin{figure}
    \centering
    \includegraphics[width=0.9\linewidth]{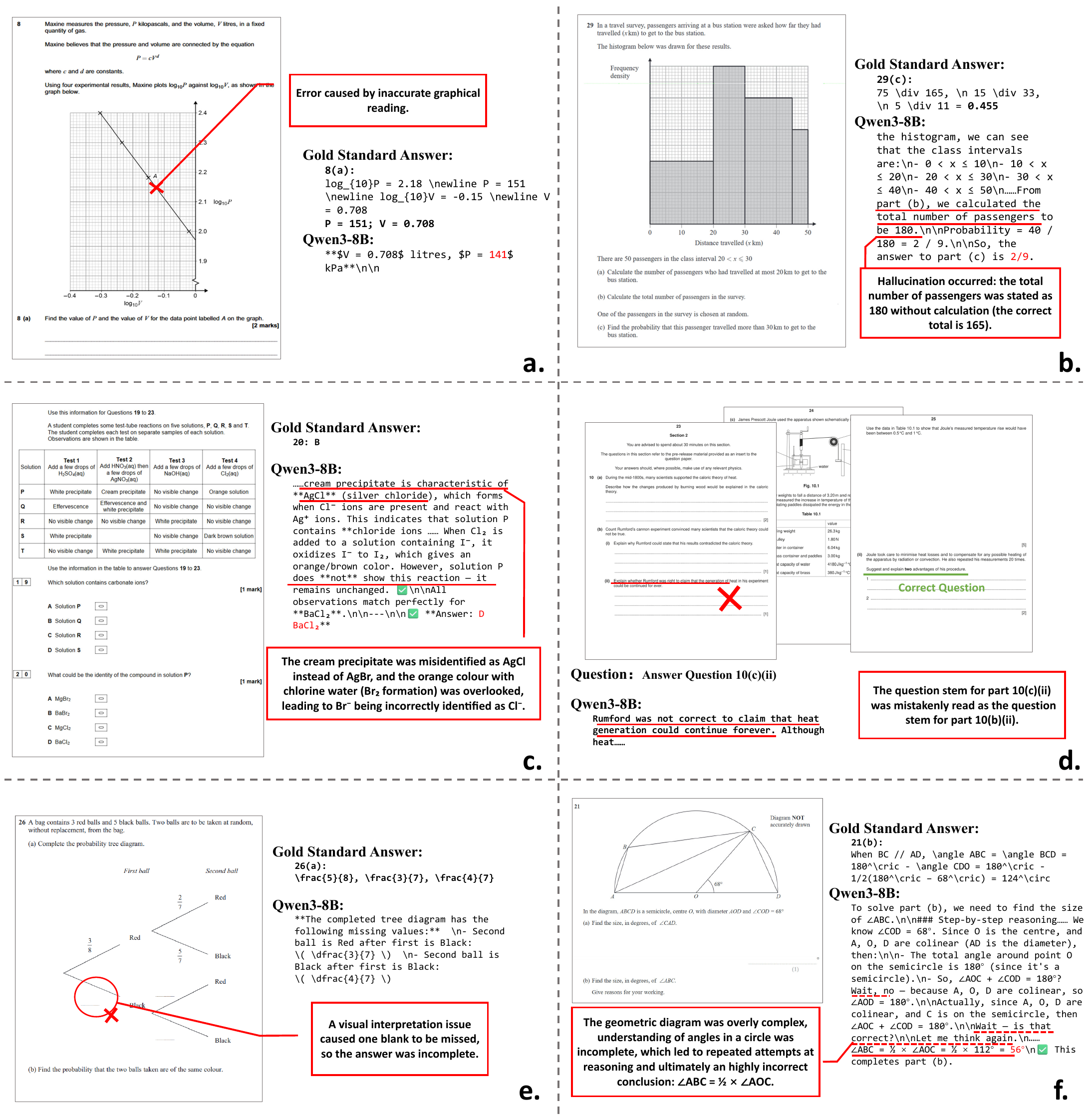}
    \caption{Examples of six representative error types.}
    \label{fig:case}
\end{figure}

Through a comprehensive error analysis, we identify several representative failure modes and present typical examples in Figure~\ref{fig:case}. For consistency, all cases are drawn from Qwen3-VL-8B under the relevant-page setting. The red boxes provide detailed diagnoses of the underlying causes. Specifically, case~\textbf{a.} shows difficulty in chart coordinate localization; case~\textbf{b.} illustrates hallucinations caused by complex statistical visualizations; case~\textbf{c.} reveals insufficient domain-specific knowledge; case~\textbf{d.} highlights query-grounding drift in cross-page retrieval; case~\textbf{e.} indicates difficulty in capturing abstract layout information; and case~\textbf{f.} shows that complex geometric diagrams can induce cyclic stalling in CoT reasoning.

These errors suggest that TestHallVQA operates not only at a variable-length, document-level scale, but also poses rich and multidimensional visual challenges.

\section{Conclusion}
In this paper, we identify the phenomenon of redundant visual contamination, provide a symbolic theoretical characterization of its underlying mechanism, and introduce TestHallVQA, a novel multi-page VQA benchmark that combines document-level redundant context with deep visual reasoning, thereby bridging the gap between existing planar VQA benchmarks. We further propose F1-R\textsuperscript{2}, a two-dimensional metric for jointly evaluating the retrieval and reasoning capabilities of LVLMs under multiple regimes of redundant visual contamination. Extensive experiments and analyses validate our theoretical findings and reveal that current LVLMs still perform unsatisfactorily in document-level settings with redundant visual inputs. We hope that our benchmark and findings will draw broader attention from the research community to document-level deep-reasoning VQA and inspire future advances in LVLMs.

\section{Acknowledgements}
We would like to thank the anonymous reviewers for their helpful comments. This work was supported by the National Natural Science Foundation of China (No.~62476066).

\clearpage
\bibliographystyle{unsrtnat}
\bibliography{references}

\clearpage
%%
%% If your work has an appendix, this is the place to put it.
\appendix
\section*{Appendix}

\tableofcontents
\addtocontents{toc}{\protect\setcounter{tocdepth}{5}}

\section{Theoretical Derivation about the impact of redundant information}
\label{theo}
To characterize how redundant information disrupts LLM reasoning, we quantify this via symbolic derivation. 

We consider two distinct scenarios. In the first, LVLMs perform multimodal reasoning and analysis within a context devoid of interference, where both the input images and textual content are solely intended to provide evidence for problem-solving. In contrast, real-world scenarios are considerably more complex: beyond useful cues, the input often contains a substantial amount of irrelevant content. At the visual level, this manifests as numerous unrelated images, where relevant images are interspersed among all images and jointly provided as input to the LLM. 

Let \(M\) denote the number of relevant tokens received by the model in the idealized setting, where the input contains only the question string and the visually relevant pages associated with the question, collectively covering all evidence required for question answering; this corresponds to a clean-case scenario. In the presence of \(N\) irrelevant visual tokens, the total number of tokens becomes \(M+N\). This redundancy serves to divert the model's attention from the relevant tokens to the irrelevant ones, thereby contaminating the hidden representations and diminishing the effectiveness of the information. 

However, it is important to emphasize that such redundant information must exhibit a certain degree of confusability. Specifically, the visual features of the irrelevant images should bear strong similarity to the target problem, thereby inducing a high level of distraction and semantic ambiguity. Only under this condition do the redundant tokens possess substantial semantic attraction; otherwise, the model will allocate negligible attention to these tokens.

We start from \textbf{the first layer}. Let the attention computation for the \(x\)-th token in the clean case be represented as \(h^1_x\). Upon the introduction of \(N\) irrelevant visual tokens, the hidden state for the \(x\)-th token after the attention layer evolves to \(h'^1_x\):
\begin{align}
    h^1_x = \hat{h}^0_{x} + & \alpha_1V_1 + \alpha_2V_2+\ldots+\alpha_MV_M, \\
    h'^1_x =\hat{h}'^0_{x} + & \beta_1V'_1 + \beta_2V'_2+\ldots+\beta_{M+N}V'_{M+N},
\end{align} 
\noindent where \(\hat{h}^0_{x}\) and \(\hat{h}'^0_{x}\) denote the output states of the previous decoder block (here, the initial embeddings), representing the effect of the residual connection, \(\alpha_i, \beta_i\) represent the attention weights, and \(V_i, V'_i \in \mathbb{R}^d\) are the \emph{value} vectors at the \(i\)-th position. The order of the tokens does not affect the derivation; for simplicity, we assume that the first \(M\) tokens are relevant, while the subsequent \(N\) tokens are irrelevant.

Due to the normalization of attention scores, both \(\sum_{i=1}^M \alpha_i\) and \(\sum_{j=1}^{M+N} \beta_j\) are equal to 1. After the addition of redundant tokens, the attention scores for the \(M\) relevant tokens are diluted and redistributed across the \(N\) irrelevant tokens. To standardize the notation, let the transformation of the average scores before and after the redistribution in relevant tokens be expressed as:
\begin{equation}
    \beta_i = \frac{M}{M+N} \cdot \delta\alpha_i, \quad i = 1,2,\ldots,M,
\label{eq:abc}
\end{equation}
\noindent where \(\delta\) is a harmonic coefficient. For convenience, we assume that the \(\delta\) value is equal across all tokens after averaging. Given that the model inherently exhibits a strong bias towards relevant information, the degree of attenuation of the attention scores for the \(M\) relevant tokens is not as severe as \(\frac{M}{M+N}\) compared to the clean case, hence \(\delta > 1\).

Therefore, the expression for \(h'_x\) in the first layer of the decoder block is given by:
 \begin{equation} \begin{split}
    h'^1_x = & \hat{h}'^0_{x} + \beta_1V'_1 + \beta_2V'_2+\ldots+\beta_{M+N}V'_{M+N} \\
    =& \hat{h}'^0_{x}+ \frac{\delta M}{M+N}(\alpha_1V'_1 + \alpha_2V'_2+\ldots+ \alpha_MV'_M) \\
     &+\beta_{M+1}V'_{M+1}+\ldots+\beta_{M+N}V'_{M+N} \\
    =& \hat{h}'^0_{x}+\frac{\delta M}{M+N}\sum^{M}_{i = 1}\alpha_iV'_i+\sum^{M+N}_{j=M+1}\beta_jV'_j, \quad \delta>1.
    \end{split}
\end{equation}

The computation of \(h'_x\) above reflects the fact that, in the presence of redundant context, after the first layer of attention, the hidden state at each token position will be mixed with information derived from the irrelevant tokens. As a result, the information purity of the hidden state is diminished.

We quantify the \textbf{information purity} of the \(x\)-th token after the first layer as \(\eta\), which represents the proportion of information from relevant tokens in the \(x\)-th hidden state after the first layer of attention, relative to the total value of the \(x\)-th hidden state:
\begin{equation}\begin{split}
    \eta =&\frac{\hat{h}'^0_{x} + \frac{\delta M}{M+N}\sum^{M}_{i = 1}\alpha_iV'_i}{\hat{h}'^0_{x}+\frac{\delta M}{M+N}\sum^{M}_{i = 1}\alpha_iV'_i+\sum^{M+N}_{j=M+1}\beta_jV'_j} \\
    =&\frac{1+\frac{\hat{h}'^0_{x}}{\frac{\delta M}{M+N}\sum^{M}_{i = 1}\alpha_iV'_i}}{1+\frac{\hat{h}'^0_{x}}{\frac{\delta M}{M+N}\sum^{M}_{i = 1}\alpha_iV'_i}+\frac{M+N}{\delta M}\cdot \frac{\sum^{M+N}_{j=M+1}\beta_jV'_j}{\sum^{M}_{i=1}\alpha_iV'_i}}\\
    =&\frac{1+\frac{\hat{h}'^0_{x}}{\sum^{M}_{i = 1}\beta_iV'_i}}{1+\frac{\hat{h}'^0_{x}}{\sum^{M}_{i = 1}\beta_iV'_i}+\frac{M+N}{\delta M}\cdot \frac{\sum^{M+N}_{j=M+1}\beta_jV'_j}{\sum^{M}_{i=1}\alpha_iV'_i}}\\
     \in& (0,1),\quad \delta >1.\\
\end{split}
\label{eq:purity-appendix}
\end{equation}
At this point, we have obtained a dilution factor \(\eta \in (0,1)\). Since \(\frac{\sum^{M+N}_{j=M+1}\beta_jV'_j}{\sum^{M}_{i=1}\alpha_iV'_i}\) represents the weighted value of the \(N\) redundant tokens divided by the weighted value of the \(M\) relevant tokens, it is primarily influenced by the ratio of \(M\) to \(N\). In the extreme case, this ratio approaches \(\frac{N}{M}\). However, due to the model's inherent resistance to redundant information, the attention weights \(\beta\) allocated to the redundant tokens will be lower than those of the relevant tokens. As a result, the numerator will be smaller than the ideal case. Moreover, the weights in the denominator correspond to the original attention weights \(\alpha\) of the \(M\) relevant tokens, rather than the diluted weights \(\beta\), which further increases the denominator. Consequently, the ratio cannot fully reach the value of \(\frac{N}{M}\). To account for this, we introduce a harmonic coefficient \(\lambda\), yielding a ratio of \(\frac{\lambda N}{M}\), where \(0 < \lambda < 1\).

Therefore, the original equation (\ref{eq:purity-appendix}) can be expressed as:
\begin{equation} \begin{split} 
    \eta &=\frac{1+\frac{\hat{h}'^0_{x}}{\sum^{M}_{i = 1}\beta_iV'_i}}{1+\frac{\hat{h}'^0_{x}}{\sum^{M}_{i = 1}\beta_iV'_i}+\frac{M+N}{\delta M}\cdot \frac{\sum^{M+N}_{j=M+1}\beta_jV'_j}{\sum^{M}_{i=1}\alpha_iV'_i}}\\
    &\Rightarrow\frac{1+\frac{\hat{h}'^0_{x}}{\sum^{M}_{i = 1}\beta_iV'_i}}{1+\frac{\hat{h}'^0_{x}}{\sum^{M}_{i = 1}\beta_iV'_i}+\frac{M+N}{\delta M} \cdot \frac{\lambda N}{M}} \\
    &\in (0,1),\quad \delta >1, \lambda \in (0,1) .
\end{split} \label{eq:14}  \end{equation}

By analogy, referring to Eqs. (\ref{eq:purity-appendix}), on the \(x\)-th token, after reaching the second layer, the ratio of the value originating from relevant information in the hidden state to the total value is given by:
\begin{equation}
    \eta_2 =\frac{(\hat{h}'^1_{x}+\frac{\delta M}{M+N}\sum^{M}_{i = 1}\alpha^2_iV'^2_i )\cdot \eta}{\hat{h}'^1_{x} +\frac{\delta M}{M+N}\sum^{M}_{i = 1}\alpha^2_iV'^2_i+\sum^{M+N}_{j=M+1}\beta^2_jV'^2_j}.
\label{eq:acc}
\end{equation}

\noindent In the formula, ``\(2\)'' denotes ``the second layer'', not ``square''. Here, we explain why each value in the numerator must be multiplied by the dilution factor \(\eta\) from the first layer: At this point, even the hidden states at the positions of the relevant tokens are contaminated by redundant information from the previous layer. Not all the information at these positions is relevant, and the proportion of relevant information has already been computed as \(\eta\). Therefore, the actual relevant information at the \(i\)-th position in this layer is \(\hat{h}'^1_i W^2_V \cdot \eta = V'^2_i \cdot \eta\), which is why we multiply each value in the numerator by \(\eta\).

Finally, analogous to Eqs. (\ref{eq:purity-appendix}) and (\ref{eq:14}) in the first layer, simplifying Eq. (\ref{eq:acc}) yields the information purity after the second layer:
\begin{equation}\begin{split}
    \eta_2 =&\frac{(\hat{h}'^1_{x}+\frac{\delta M}{M+N}\sum^{M}_{i = 1}\alpha^2_iV'^2_i )\cdot \eta}{\hat{h}'^1_{x} +\frac{\delta M}{M+N}\sum^{M}_{i = 1}\alpha^2_iV'^2_i+\sum^{M+N}_{j=M+1}\beta^2_jV'^2_j} \\
    \Rightarrow&\frac{1+\frac{\hat{h}'^1_{x}}{\sum^{M}_{i = 1}\beta^2_iV'^2_i}}{1+\frac{\hat{h}'^1_{x}}{\sum^{M}_{i = 1}\beta^2_iV'^2_i}+\frac{M+N}{\delta M} \cdot \frac{\lambda N}{M}} \cdot \eta\\
    =&\frac{1+\frac{\hat{h}'^0_{x}\cdot \eta  }{\sum^{M}_{i = 1}\beta_iV'_i \cdot \eta}}{1+\frac{\hat{h}'^0_{x}\cdot \eta}{\sum^{M}_{i = 1}\beta_iV'_i \cdot \eta}+\frac{M+N}{\delta M} \cdot \frac{\lambda N}{M}} \cdot \eta\\
    =&\frac{1+\frac{\hat{h}'^0_{x}}{\sum^{M}_{i = 1}\beta_iV'_i}}{1+\frac{\hat{h}'^0_{x}}{\sum^{M}_{i = 1}\beta_iV'_i}+\frac{M+N}{\delta M} \cdot \frac{\lambda N}{M}}\cdot \eta
 = \eta^2 \in (0,1).
\end{split}\end{equation}
In other words, after each decoder block, the proportion of relevant information in the hidden state (the information purity) is scaled by \(\eta\). Consequently, for an \(L\)-layer decoder, the effective information retained in the final hidden state is only \(\eta^L\) of its original value, leading to an exponential compression of the informative content:
\begin{equation}
    \eta_{_L} = \eta^L, \quad L\in \mathbb{Z}.
\end{equation}

Furthermore, we analyze the influence of $M$ and $N$ on $\eta$: 

From the result in Eq.~(\ref{eq:14}), we can observe that the term
\(
1 + \frac{\hat{h}'^{0}_{x}}{\sum_{i=1}^{M} \beta_i V'_i}
\)
appears entirely in the denominator and constitutes a constant greater than $1$. For notational simplicity, we denote this constant by $b$. We focus on the remaining denominator term\(\frac{M+N}{\delta M} \cdot \frac{\lambda N}{M}\):
\begin{equation}
\begin{split}
\eta 
&\Rightarrow \frac{1+\frac{\hat{h}'^0_{x}}{\sum^{M}_{i = 1}\beta_iV'_i}}{1+\frac{\hat{h}'^0_{x}}{\sum^{M}_{i = 1}\beta_iV'_i}+\frac{M+N}{\delta M} \cdot \frac{\lambda N}{M}} \\
&= \frac{b}{b + \frac{M+N}{\delta M} \cdot \frac{\lambda N}{M}} = \frac{b}{b + \frac{\lambda MN + \lambda N^2}{\delta M^2}} = \frac{b}{b + \frac{\lambda \frac{N}{M} + \lambda \left(\frac{N}{M}\right)^2}{\delta}} \\
&= \frac{b}{b + \frac{\lambda}{\delta}
\left(\frac{N}{M} + \left(\frac{N}{M}\right)^2\right)} .
\end{split}
\end{equation}

As shown above, the denominator reduces to a univariate quadratic form $x^2 + x$, where $x = \frac{N}{M}$. Since $\frac{N}{M} > 0$ and the function $x^2 + x$ is monotonically increasing for $x > 0$, the denominator increases monotonically as $\frac{N}{M}$ grows, i.e., as redundant tokens constitute a larger proportion of the input. Consequently, $\eta$ decreases monotonically, indicating a degradation in information purity.

We therefore conclude that as the ratio of redundant tokens to relevant tokens increases, the purity of information in the hidden states decreases, which aligns with intuitive expectations.

Even if \(\eta\) is a value close to \(1\), its impact under exponentiation can still be substantial. Given the increasing number of layers in modern LLMs and vision encoders, this phenomenon warrants more attention. The contamination of the hidden state with redundant information as additional tokens are incorporated hampers the ability to capture useful signals, ultimately disrupting both vocabulary decoding and output generation.

Given the inevitability of redundancy in real-world data, we argue that an LVLM's capacity to maintain reasoning performance under redundant conditions---rather than under an unrealistically clean setting---should be given greater emphasis as an extended evaluation criterion. Our benchmark is designed precisely for this setting.

\clearpage
\section{Multi-Level Redundant Context Injection Procedure}
\label{sec:regimes}

In TestHallVQA, each question is associated with multiple levels of contextual redundancy, which we refer to as \textbf{regimes}. Under our design, regimes are primarily distinguished by the number of pages they include, with an approximate interval of ten pages between adjacent regimes. Based on dataset statistics, the maximum length of exam documents in our data source is 44 pages. Accordingly, we define up to five regimes (\emph{Regime 1} through \emph{Regime 5}). The annotation procedure is described as Algorithm \ref{alg:redundant_context}:

\begin{algorithm}[ht]
\caption{Multi-Level Redundant Context Generation}
\label{alg:redundant_context}
\begin{algorithmic}[1]
\Require Document total pages $N$, Relevant page set $R$ ($|R| \leq 4$)
\Ensure $R\leq N$
\State $\hat{N} \gets 10 \times \text{round}(N / 10)$ \Comment{Round maximum pages to nearest multiple of 10}
\State $K \gets \hat{N} / 10 + 1$ \Comment{Total number of regimes}
\State Initialize an array $P$ of size $K$ to store page counts for each regime
\State $P_1 \gets |R|$ \Comment{Regime 1 page count is the size of the relevant page set}
\For{$i = 2$ \textbf{to} $K$}
    \State $P_i \gets 10 \times (i - 1)$ \Comment{Regime pages: 10, 20, 30, ...}
\EndFor
\If{$N < \hat{N}$}
    \State $P_K \gets N$ \Comment{Restore the last regime's page count to $N$}
\EndIf
\State $\mathcal{S} \gets \emptyset$ \Comment{The set of redundant page collections (Regimes)}
\For{$i = 1$ \textbf{to} $K$}
    \State $S_i \gets R$
    \State \textit{// Step 1: Fill internal gaps between min and max pages}
    \State $p_{min} \gets \min(R)$
    \State $p_{\max} \gets \max(R)$
    \State $C \gets \{p \mid p_{min} \leq p \leq p_{\max}\} \setminus S_i$
    \While{$|S_i| < P_i$ \textbf{and} $C \neq \emptyset$}
        \State Extract a page $p$ from $C$ and $S_i \gets S_i \cup \{p\}$
    \EndWhile
    \State \textit{// Step 2: Automated left-right expansion strategy}
    \State $left \gets p_{min} - 1$
    \State $right \gets p_{\max} + 1$
    \While{$|S_i| < P_i$ \textbf{and} ($left \geq 1$ \textbf{or} $right \leq N$)}
        \State $dir \gets \text{RandomChoice}(\{\text{left}, \text{right}\})$
        \If{$dir = \text{left}$ \textbf{and} $left \geq 1$}
            \State $S_i \gets S_i \cup \{left\}$
            \State $left \gets left - 1$
        \ElsIf{$dir = \text{right}$ \textbf{and} $right \leq N$}
            \State $S_i \gets S_i \cup \{right\}$
            \State $right \gets right + 1$
        \EndIf
    \EndWhile
    \State $\mathcal{S} \gets \mathcal{S} \cup \{S_i\}$
\EndFor
\State \Return $\mathcal{S}$
\end{algorithmic}
\end{algorithm}

Through this annotation process, the difference in page count between adjacent redundant regimes is at least five pages (except for \emph{Regime 1}, which represents strictly relevant pages). This design ensures clear separability between redundancy levels while maximizing the utilization of available exam pages.

\clearpage
\section{TestHallVQA Dataset Details}
\label{sec:stati}

Table \ref{tab:static} presents key statistical information about TestHallVQA, enabling researchers to gain an intuitive overview of the dataset's composition. Table \ref{tab:static2} reports results on a \emph{mini-test} set, which is obtained via class-wise random sampling from the full TestHallVQA dataset. This subset preserves the distribution of problem types while significantly reducing evaluation cost, enabling efficient comparison and ablation analysis without affecting the overall evaluation trends.

\begin{table}[h]
    \centering

    % ==================== Full test set ====================
    \begin{minipage}[t]{0.485\textwidth}
        \vspace{0pt}
        \centering
        \setlength{\tabcolsep}{3pt}
        \renewcommand{\arraystretch}{0.95}

        \begin{tabularx}{\linewidth}{
            >{\raggedright\arraybackslash}X
            >{\raggedleft\arraybackslash}p{1.25cm}
        }
        \toprule
        \textbf{Statistic} & \textbf{Number} \\
        \midrule

        \multicolumn{2}{l}{\textbf{Total}} \\
        \quad Total Question ID Samples       & 10,242 \\
        \quad Total Subquestion Samples       & 11,317 \\
        \quad Total Images                    & 7,155  \\

        \midrule
        \multicolumn{2}{l}{\textbf{Subjects}} \\
        \quad Mathematics                     & 3,302 \\
        \quad Physics                         & 3,231 \\
        \quad Chemistry                       & 3,709 \\

        \midrule
        \multicolumn{2}{l}{\textbf{Question Types}} \\
        \quad Normal                          & 4,189 \\
        \quad Calculate                       & 2,755 \\
        \quad Statement                       & 1,940 \\
        \quad Choice                          & 1,923 \\
        \quad Proof                           & 410   \\
        \quad Binary                          & 100   \\

        \midrule
        \multicolumn{2}{l}{\textbf{Context Regimes}} \\
        \quad Regime 1 (relevant pages, 1--4 pages)
                                              & 10,242 \\
        \quad Regime 2 (5--10 pages)           & 9,327  \\
        \quad Regime 3 (15--20 pages)          & 7,977  \\
        \quad Regime 4 (25--30 pages)          & 3,772  \\
        \quad Regime 5 (35--40 pages)          & 859    \\

        \midrule
        \multicolumn{2}{l}{\textbf{Maximum, Minimum, and Average}} \\
        \quad Maximum Exam Document Length    & 44    \\
        \quad Minimum Exam Document Length    & 1     \\
        \quad Average Exam Document Length    & 18.73 \\
        \quad Average Regime 1 Pages           & 1.55  \\
        \quad Average Regime 2 Pages           & 9.92  \\
        \quad Average Regime 3 Pages           & 19.22 \\
        \quad Average Regime 4 Pages           & 28.62 \\
        \quad Average Regime 5 Pages           & 37.13 \\
        \bottomrule
        \end{tabularx}

        \captionof{table}{
            Detailed statistics of the full TestHallVQA test set.
            Some question IDs contain multiple subquestions, each of
            which requires a separate answer.
        }
        \label{tab:static}
    \end{minipage}
    \hfill
    % ==================== Mini-test set ====================
    \begin{minipage}[t]{0.485\textwidth}
        \vspace{0pt}
        \centering
        \setlength{\tabcolsep}{3pt}
        \renewcommand{\arraystretch}{0.95}

        \begin{tabularx}{\linewidth}{
            >{\raggedright\arraybackslash}X
            >{\raggedleft\arraybackslash}p{1.25cm}
        }
        \toprule
        \textbf{Statistic} & \textbf{Number} \\
        \midrule

        \multicolumn{2}{l}{\textbf{Total}} \\
        \quad Total Question ID Samples       & 1,000 \\
        \quad Total Subquestion Samples       & 1,133 \\
        \quad Total Images                    & 734   \\

        \midrule
        \multicolumn{2}{l}{\textbf{Subjects}} \\
        \quad Mathematics                     & 297 \\
        \quad Physics                         & 345 \\
        \quad Chemistry                       & 358 \\

        \midrule
        \multicolumn{2}{l}{\textbf{Question Types}} \\
        \quad Normal                          & 410 \\
        \quad Calculate                       & 269 \\
        \quad Statement                       & 198 \\
        \quad Choice                          & 201 \\
        \quad Proof                           & 43  \\
        \quad Binary                          & 12  \\

        \midrule
        \multicolumn{2}{l}{\textbf{Context Regimes}} \\
        \quad Regime 1 (relevant pages, 1--4 pages)
                                              & 1,000 \\
        \quad Regime 2 (5--10 pages)           & 926   \\
        \quad Regime 3 (15--20 pages)          & 781   \\
        \quad Regime 4 (25--30 pages)          & 373   \\
        \quad Regime 5 (35--40 pages)          & 88    \\

        \midrule
        \multicolumn{2}{l}{\textbf{Maximum, Minimum, and Average}} \\
        \quad Maximum Exam Document Length    & 41    \\
        \quad Minimum Exam Document Length    & 1     \\
        \quad Average Exam Document Length    & 19.56 \\
        \quad Average Regime 1 Pages           & 1.41  \\
        \quad Average Regime 2 Pages           & 9.35  \\
        \quad Average Regime 3 Pages           & 19.72 \\
        \quad Average Regime 4 Pages           & 29.03 \\
        \quad Average Regime 5 Pages           & 36.50 \\
        \bottomrule
        \end{tabularx}

        \captionof{table}{
            Detailed statistics of the TestHallVQA mini-test set.
        }
        \label{tab:static2}
    \end{minipage}
\end{table}

\clearpage
\section{Ablation Setup for the \texorpdfstring{$S_{RT}$}{SRT} Metric}
\label{sec:SRT}

As mentioned in the main text, to demonstrate the advantage of $S_{RT}$ in F1-R\textsuperscript{2}, we compare it with several baseline metrics, including step retention ratio (SRR), end-to-end retention (E2E), normalized AUC (N-AUC), and log-based decay (LD). In this section, we present the computation of each baseline metric in detail.

We continue to use the illustrative example from Section \ref{sec:F1R2}. Specifically, for a sub-test set with three regimes, let the average accuracies under Regime 1 to Regime 3 be denoted by $e_1, e_2, e_3 \in [0,100]$, respectively.

\textbf{Step retention ratio (SRR)} directly characterizes performance degradation using relative decay, namely, the score ratio between two consecutive regimes, i.e., $\frac{e_2}{e_1}$ and $\frac{e_3}{e_2}$. After normalization, it is defined as:
\begin{equation}
    S_{RT\text{-}SRR}=\frac{\Delta p_1 \cdot \frac{e_2}{e_1}+\Delta p_2 \cdot \frac{e_3}{e_2}}{\Delta p_1+\Delta p_2},
\end{equation}
where $\Delta p_1=\bar p_2-\bar p_1$ and $\Delta p_2=\bar p_3-\bar p_2$. The interpretation of this metric is straightforward: it measures how much performance is retained for each additional block of redundant pages.

\textbf{End-to-end retention (E2E)} measures how much performance remains from the oracle setting to the most challenging regime. It can be viewed as a simplified version of step retention ratio, with particularly strong interpretability. Instead of considering the intermediate trend of the performance curve, it directly compares the initial and final regimes:
\begin{equation}
    S_{RT\text{-}E2E}=\frac{e_3}{e_1}.
\end{equation}

\textbf{Normalized AUC (N-AUC)} is closely related to the rotation-angle formulation, but removes the $\arctan$ transformation. It is therefore well suited to addressing a central question: does the angle transformation truly provide additional discriminative power, or is the area-based characterization alone already sufficient? 

Since the rotation-angle metric in the main paper is itself motivated by the interpretation of the ``removed triangular area,'' the most natural baseline is to directly use the ratio between the trapezoidal area remaining after removing the triangle and the area of the original rectangle:
\begin{equation}
S_{\text{RT-AUC}}=
\frac{\frac{e_1+e_2}{2}\Delta p_1}{e_1\Delta p_1} \cdot \frac{\Delta p_1}{\Delta p_1 + \Delta p_2}
+\frac{\frac{e_2+e_3}{2}\Delta p_2}{e_2\Delta p_2} \cdot \frac{\Delta p_2}{\Delta p_1 + \Delta p_2},
\end{equation}
where $\Delta p_1=\bar p_2-\bar p_1$ and $\Delta p_2=\bar p_3-\bar p_2$. This formulation likewise yields a score in the range $[0,1]$.

\textbf{Log-Based Decay (LD)} quantifies performance degradation using a logarithmic transformation. Since $\frac{e_2}{e_1} \in [0,1]$ represents the proportion of performance retained, $1-\frac{e_2}{e_1} \in [0,1]$ correspondingly measures the proportion of performance lost. A larger loss should lead to a lower final score. The negative logarithm, i.e., $f(x)=-\log(x)$, is well suited to this purpose over the interval $[0,1]$. Moreover, it is more sensitive to values approaching zero, thereby improving sensitivity to mild degradation:
\begin{equation}
S_{\text{RT-LD}_1}^{\prime}=
-\log\left(1-\frac{e_2}{e_1} + \varepsilon\right),
\end{equation}
where $\varepsilon$ is a small constant introduced to avoid numerical overflow at zero; in our experiments, we set $\varepsilon=10^{-3}$.

We then normalize this quantity by considering its boundary cases. When $e_2=0$, corresponding to the worst possible performance, the value becomes $-\log(1+\varepsilon)$. When $e_2=e_1$, corresponding to perfect retention, the value becomes $-\log(\varepsilon)$. Therefore, the normalized score for the first segment is:
\begin{equation}
        S_{\text{RT-LD}_1}=\frac{-\log\left(1-\frac{e_2}{e_1} + \varepsilon\right)-\left(-\log(1 + \varepsilon)\right)}{-\log(\varepsilon)-\left(-\log(1 + \varepsilon)\right)} =\frac{\log(1 + \varepsilon)-\log\left(1-\frac{e_2}{e_1} + \varepsilon\right)}{\log(1 + \varepsilon)-\log(\varepsilon)}.
\end{equation}

Similarly, the score for the second segment is:
\begin{equation}
    S_{\text{RT-LD}_2} = \frac{\log(1 + \varepsilon)-\log\left(1-\frac{e_3}{e_2} + \varepsilon\right)}{\log(1 + \varepsilon)-\log(\varepsilon)}.
\end{equation}

The final normalized metric is computed as:
\begin{equation}
    S_{\text{RT-LD}} = S_{\text{RT-LD}_1}\cdot \frac{\Delta p_1}{\Delta p_1 + \Delta p_2}+S_{\text{RT-LD}_2}\cdot \frac{\Delta p_2}{\Delta p_1 + \Delta p_2}.
\end{equation}

\newpage

\section{Extended Ablation Analyses}
\label{sec:extend}

\subsection{Cross-Page Question Distractiveness Analysis}

\begin{figure}[htbp]
    \centering
    \includegraphics[width=1\linewidth]{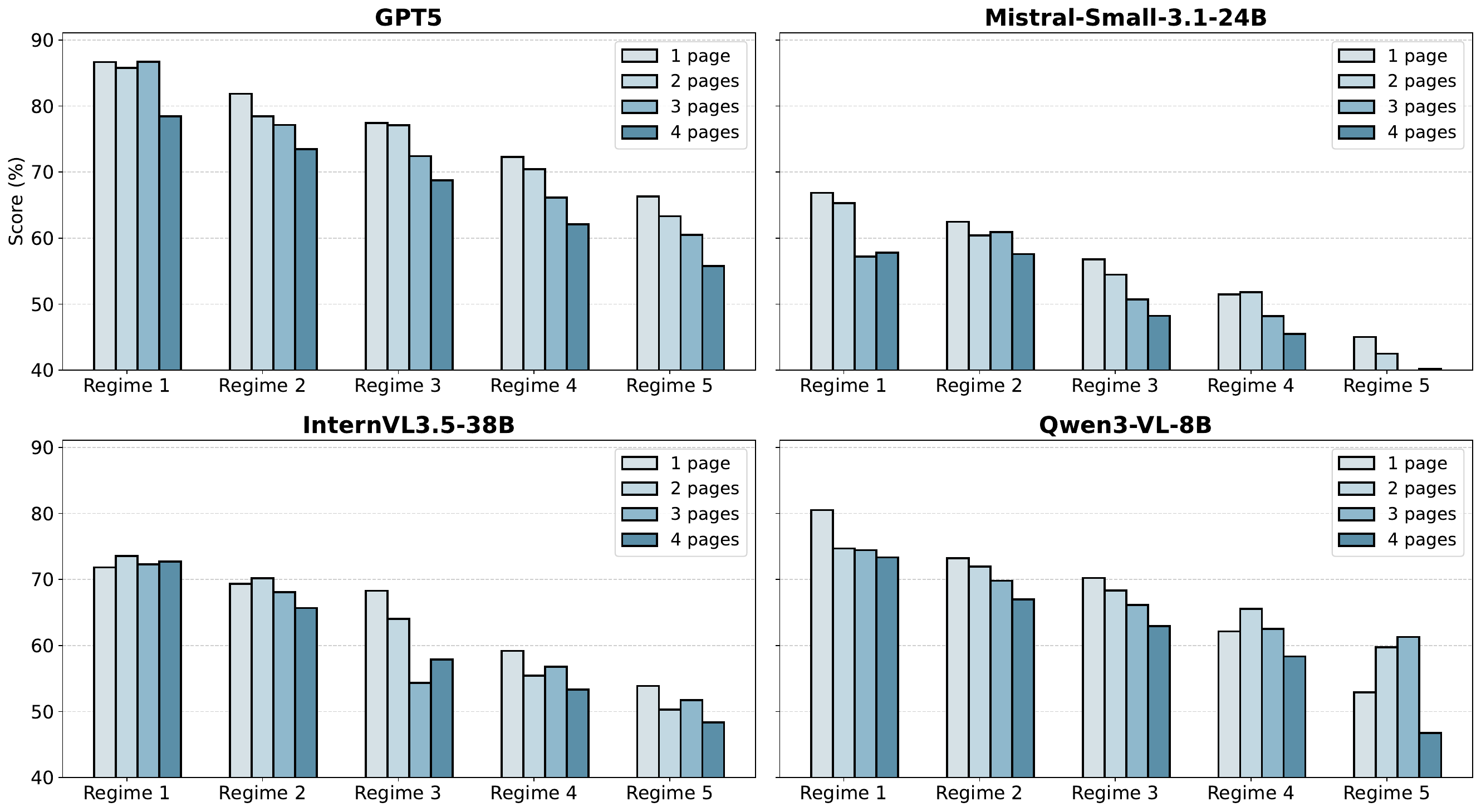}
    \caption{Performance differences for questions spanning different numbers of relevant pages.}
    \label{fig:tiao_main}
\end{figure}

As TestHallVQA involves cross-page evidence aggregation (as shown in Figure~\ref{fig:cross}), we report the accuracy on questions with different page spans in the Regime~1 subset. Figure~\ref{fig:tiao_main} shows that accuracy generally declines as the number of spanned pages increases, although the trend is not strictly monotonic. This is because, while cross-page reasoning introduces greater visual challenges, the intrinsic difficulty of the questions also plays a substantial role, and the reasoning difficulty is only weakly correlated with the number of pages spanned.

\subsection{Redundant Image Distractiveness Analysis}
\begin{figure}[htbp]
    \centering
    \includegraphics[width=\linewidth]{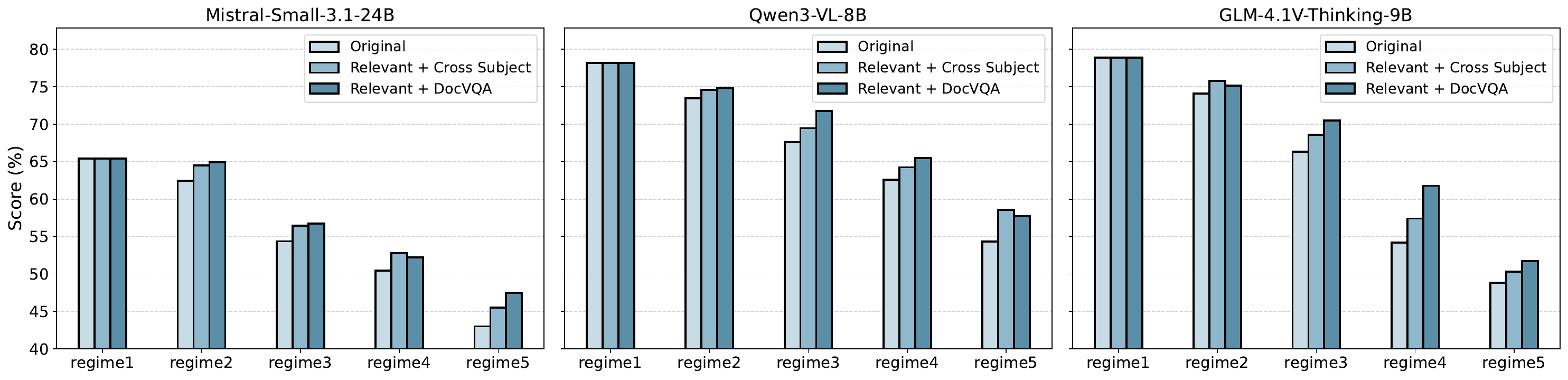}
    \caption{Comparison of Model Performance Perturbations under Different Redundant Image Selections.}
    \label{fig:tiao}
\end{figure}
To examine how the distractiveness of redundant information in TestHallVQA affects model performance, we replace the original redundant images, which are drawn from the same exam paper as the relevant image, with two alternative sources: (1) exam papers from different subjects within TestHallVQA, and (2) DocVQA \cite{docvqa}.

For cross-subject redundancy, we ensure that question index ranges do not overlap. Specifically, for each question ID, we randomly select an exam paper containing the same ID, identify the page range of its relevant context, and sample redundant pages only from outside this range via random left--right expansion. For DocVQA-based redundancy, pages are randomly sampled from DocVQA until the target number of redundant images is reached. Compared with exam paper images, DocVQA pages are much denser in text and visually distinct from the original exam-style documents.

In all settings, redundant images are resized to match the resolution of the corresponding relevant images, and the number of images is kept identical to the original annotations, ensuring consistent token length across configurations.

We annotate these settings on the mini-test split of TestHallVQA and evaluate several stable models from Table~\ref{tab:main}. To isolate model behavior under different context scales, we report accuracy for each regime separately, without computing cross-regime F1-R\textsuperscript{2}. As shown in the figure \ref{fig:tiao}, unstructured redundant images are less misleading to models. This suggests that existing datasets cannot achieve controlled inclusion of highly misleading redundant visual context through data augmentation alone, further underscoring the unique value of TestHallVQA.

\section{Limits of LVLMs in Multi-Image Visual Encoding}
\label{sec:limit}

\begin{figure}[t]
    \centering
    \begin{subfigure}[t]{0.48\textwidth}
        \centering
        \includegraphics[width=\linewidth]{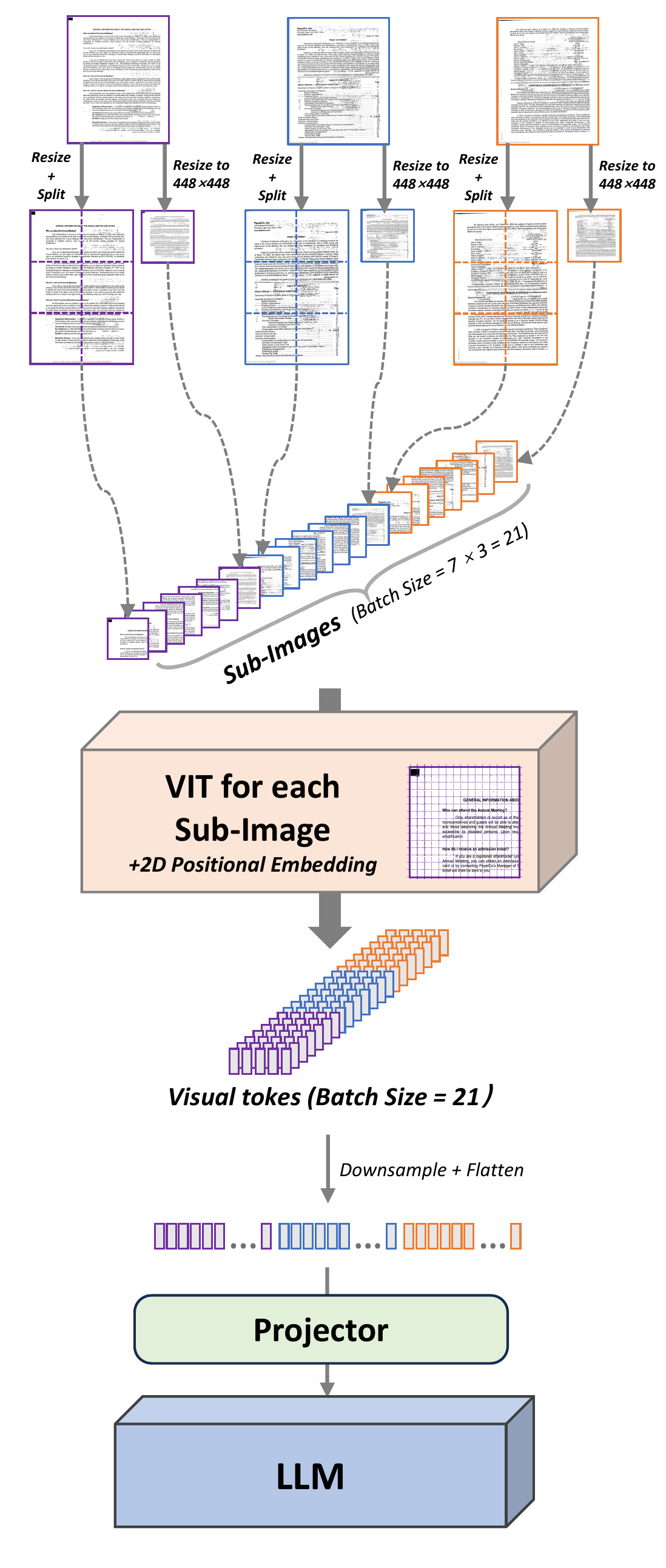}
        \caption{Visual encoding pipeline of conventional LVLMs.}
        \label{fig:vit_intern}
    \end{subfigure}
    \hfill
    \begin{subfigure}[t]{0.48\textwidth}
        \centering
        \includegraphics[width=1\linewidth]{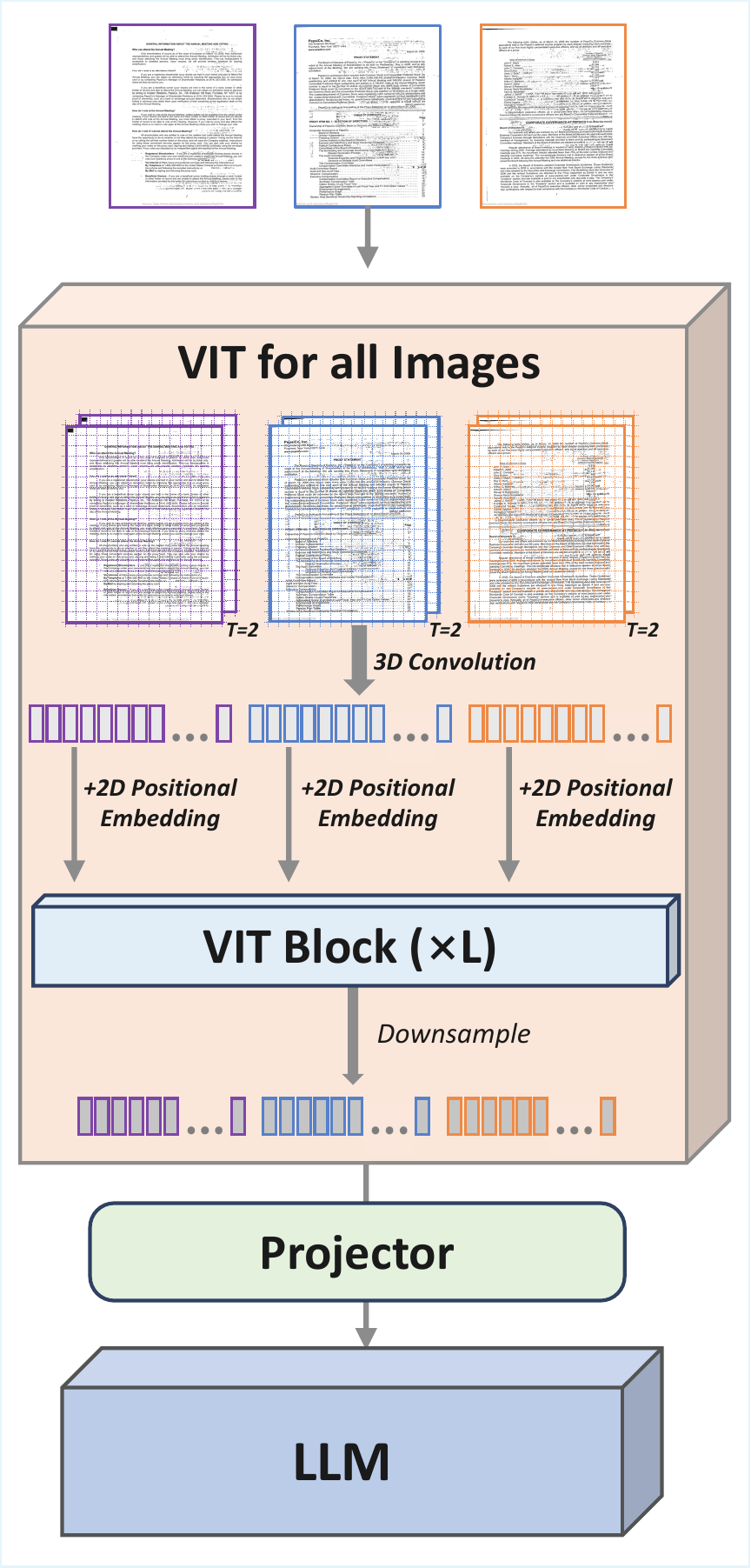}
        \caption{Visual encoding pipeline of the Qwen2.5/3-VL series.}
        \label{fig:vit_qwen}
    \end{subfigure}
    \caption{Visual encoding pipelines.}
\end{figure}

Admittedly, the fundamental cause of the degradation in LVLMs' reasoning performance under document-level contexts lies in the phenomenon of redundant visual contamination. However, through an analysis of failure cases, we observe that models frequently exhibit errors in locating the correct question identifier, corresponding to type \textbf{d.} in Figure~\ref{fig:case}. We attribute this issue to the fact that question identifiers in documents are often distributed across multiple pages, with different parts of a complete identifier scattered over several pages and organized in a sequential order. Based on this observation, we infer that LVLMs perform poorly when processing document images with inherent temporal dependencies.

Through a careful examination of both the source code and the corresponding papers, we indeed identified evidence supporting this hypothesis. We find that, when processing multiple images, the visual encoder in LVLMs is unable to capture the sequential dependencies across images; that is, throughout the visual encoding stage, different images are either treated as order-agnostic or are effectively isolated from one another. Only when the visual tokens are first linearly serialized and fed into the LLM decoder do the cross-page temporal dependencies begin to be modeled, by which point the opportunity to encode such relationships at the visual stage has already been missed.

Specifically, we analyze this issue from the perspective of LVLM architecture and summarize two of the most prevalent visual encoder paradigms adopted by contemporary LVLMs.

\subsection{Conventional LVLM Architectures}

As illustrated in Figure~\ref{fig:vit_intern}, conventional LVLMs, such as InternVL and LLaVA, are built upon the standard Vision Transformer (ViT) architecture. Input images are resized to a fixed resolution (e.g., 448) and fed into the ViT to obtain the corresponding visual tokens. When the input resolution is excessively high, a \emph{tiling} strategy is adopted: the image is divided into multiple sub-images, which are processed as a batch by the ViT, while a resized version of the original image is additionally retained as a global-context image. This procedure naturally extends to the multi-image setting, where each image is processed independently in the same manner, resulting in a larger batch of sub-images.

The limitation of this design is that it implicitly treats each sub-image as an independent image, even though each of them represents only a local region of the original image. The 2D rotary positional encoding (\textbf{2D-RoPE}) in the ViT is applied only within each individual sub-image and does not explicitly encode the spatial relationships among sub-images. As a result, sub-images are encoded in a mutually isolated manner during visual encoding and remain unaware of one another. Different sub-images do not interact until they are passed to the LLM. At that stage, the LLM applies one-dimensional RoPE to both visual and textual tokens to impose a sequential ordering.

However, \cite{meiyong} show that the norm of visual embeddings is typically one to three orders of magnitude larger than that of textual embeddings. Consequently, when positional information is introduced only at the LLM stage, the resulting positional signal is inherently weak, as it is easily dominated by the much stronger visual features.

Moreover, because inter-sub-image relationships are not modeled during visual encoding, the LLM is forced to reconstruct such dependencies from scratch. This additional burden interferes with semantic understanding and reasoning, thereby distracting the LLM from the downstream task.

\subsection{Emerging Architectures: Qwen2.5/3-VL}

The Qwen2.5/3-VL family adopts a more recent visual-encoder design, as illustrated in Figure~\ref{fig:vit_qwen}. Unlike approaches that partition an image into multiple sub-images and encode them separately, Qwen2.5/3-VL processes all visual tokens belonging to the same image jointly within the vision transformer. This allows tokens from different spatial regions of an individual image to interact directly during visual encoding.

For multi-image inputs, however, the images remain isolated from one another inside the visual encoder. Although the token sequences of multiple images may be packed together for efficient implementation, an attention mask prevents visual tokens from one image from attending to tokens belonging to another image. Therefore, the vision encoder does not establish cross-image correspondences or model the sequential relationships among images, and each image constitutes an independent attention unit within the visual-encoding stage.

The positional encodings used by the visual encoder are likewise defined independently for each image according to its two-dimensional spatial structure. They encode the height and width coordinates of tokens within an image, but do not represent the relative order of different images. For example, tokens located at the same spatial coordinates in two identically sized images may receive the same two-dimensional positional indices. This does not cause them to interact within the vision encoder, because cross-image attention is masked, but it also means that the visual encoder itself cannot distinguish or exploit the temporal ordering of the images.

After visual encoding, the resulting image-token sequences are projected and inserted into the input sequence of the LLM decoder. At this stage, Qwen-VL applies M-RoPE \cite{qwen2.5vl,Qwen3-VL}, which incorporates temporal, height, and width positional components. The temporal component assigns distinct positional indices to different images or video frames, thereby introducing their sequential order into the language-modeling stage. Consequently, cross-image relationships are modeled primarily by the LLM decoder rather than by the visual encoder. This design preserves efficient per-image visual encoding, but places the burden of integrating evidence and reasoning across multiple images on the LLM.

\subsection{Summary}
In summary, existing LVLMs are \textbf{unable to perform joint multi-image analysis} at an early stage (i.e., during visual encoding). As a result, inter-image relationships are \textbf{insufficiently modeled}, and visual content is passed to the LLM before being adequately understood. Consequently, the LLM decoder must devote additional capacity to processing and inferring temporal relationships among cross-image tokens from scratch, thereby distributing attention more diffusely and further \textbf{exacerbating the phenomenon of redundant visual contamination}.

This limitation is reflected in common error cases observed in TestHallVQA, such as misalignment in multi-level question indexing and incomplete extraction of visual elements and regions. Consequently, we expect future work to address this limitation and propose corresponding improvements.

\clearpage

\section{LLM-as-a-Judge Scoring Details}
\label{sec:judge}

We iteratively designed and refined a set of prompts---guided by
continuous feedback from LLM outputs---to ultimately obtain a robust,
fully instructed prompting framework. This framework provides
comprehensive reasoning guidance and complete logical coverage,
ensuring strong generalization across the TestHallVQA benchmark.
It is used to elicit the analytical capabilities of an LLM acting as
an evaluator, enabling it to assess the outputs of various LVLMs.

For the six problem types, we design specialized prompts tailored to
the unique reasoning requirements of each category. In this section,
we present the exact prompt templates and provide several representative
interaction examples, offering an intuitive illustration of the
evaluation process.

\begin{figure}[h]
    \centering

    \begin{subfigure}[t]{0.485\linewidth}
        \centering
        \includegraphics[width=\linewidth]
        {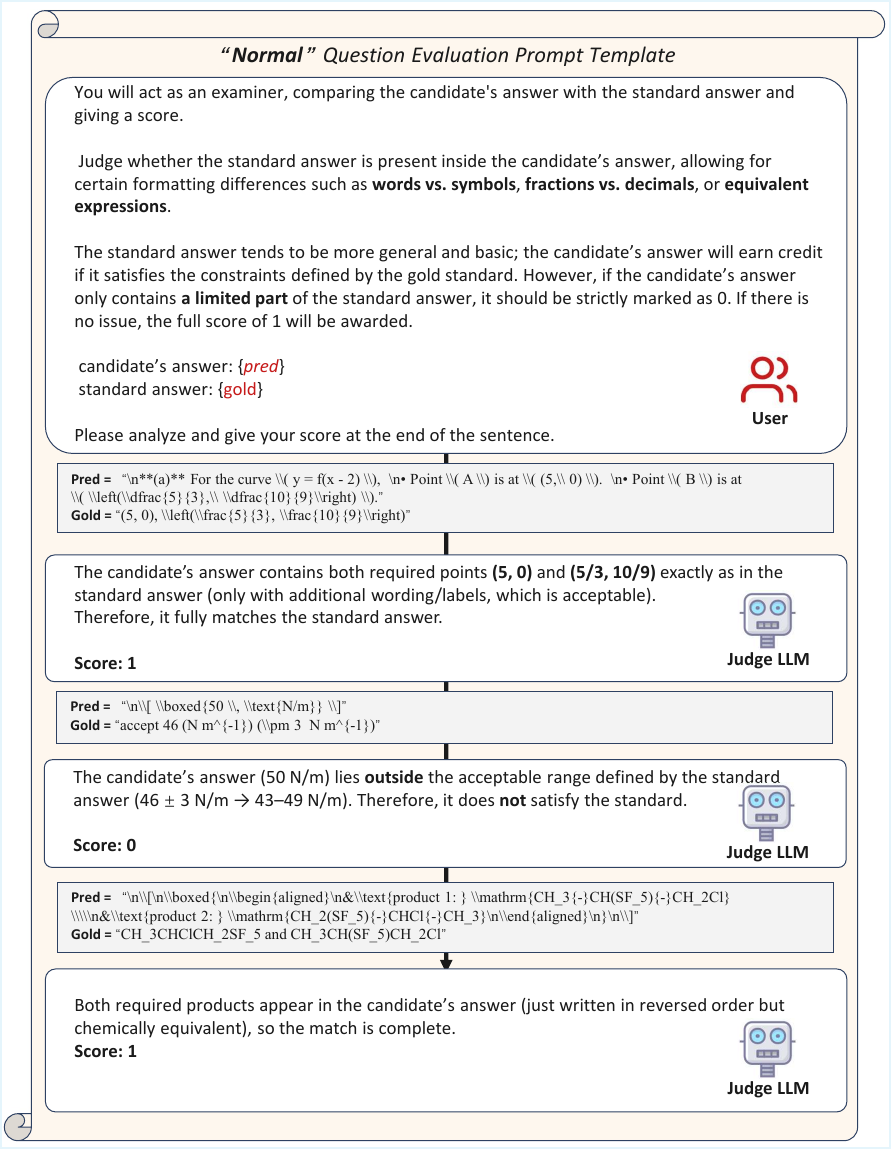}
        \caption{Scoring process for \textbf{Normal}-type questions.}
        \label{fig:normal}
    \end{subfigure}
    \hfill
    \begin{subfigure}[t]{0.485\linewidth}
        \centering
        \includegraphics[width=\linewidth]
        {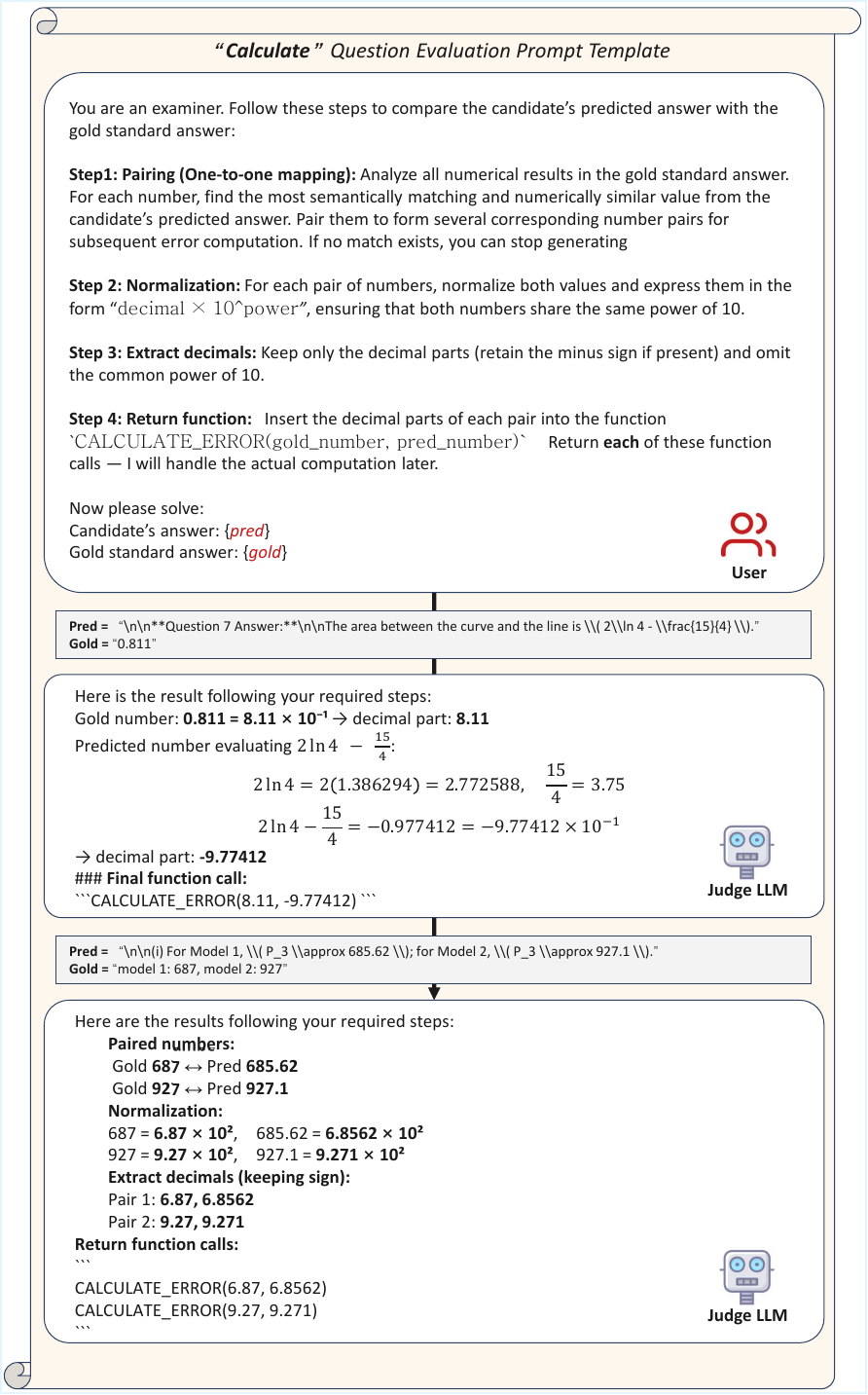}
        \caption{Scoring process for \textbf{Calculate}-type questions.}
        \label{fig:calculate}
    \end{subfigure}

    \caption{LLM-as-a-Judge scoring processes for Normal and
    Calculate questions.}
    \label{fig:judge_normal_calculate}
\end{figure}

\begin{figure}[p]
    \centering

    \begin{subfigure}[t]{0.485\linewidth}
        \centering
        \includegraphics[width=\linewidth]
        {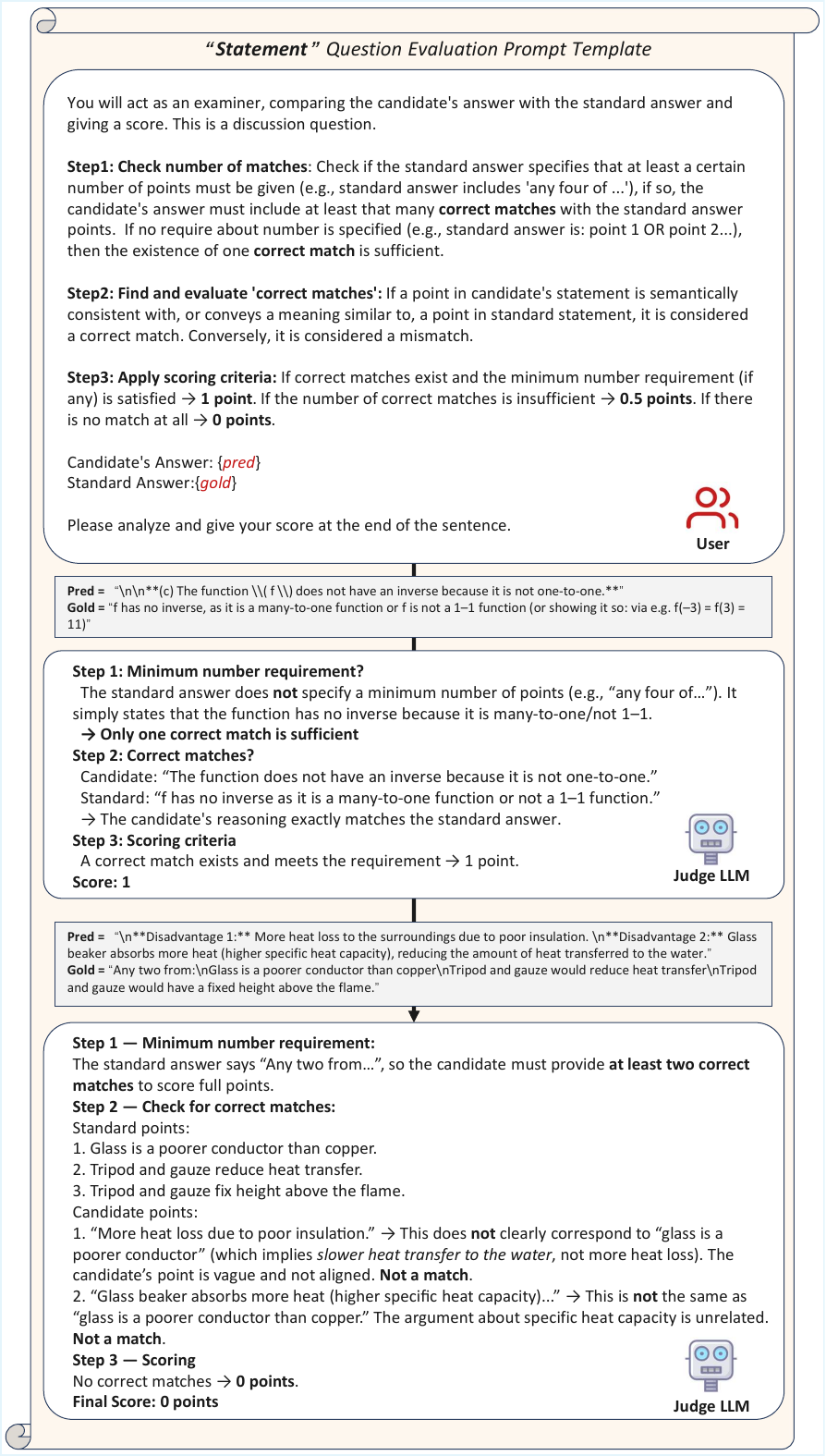}
        \caption{Scoring process for \textbf{Statement}-type questions.}
        \label{fig:statement}
    \end{subfigure}
    \hfill
    \begin{subfigure}[t]{0.485\linewidth}
        \centering
        \includegraphics[width=\linewidth]
        {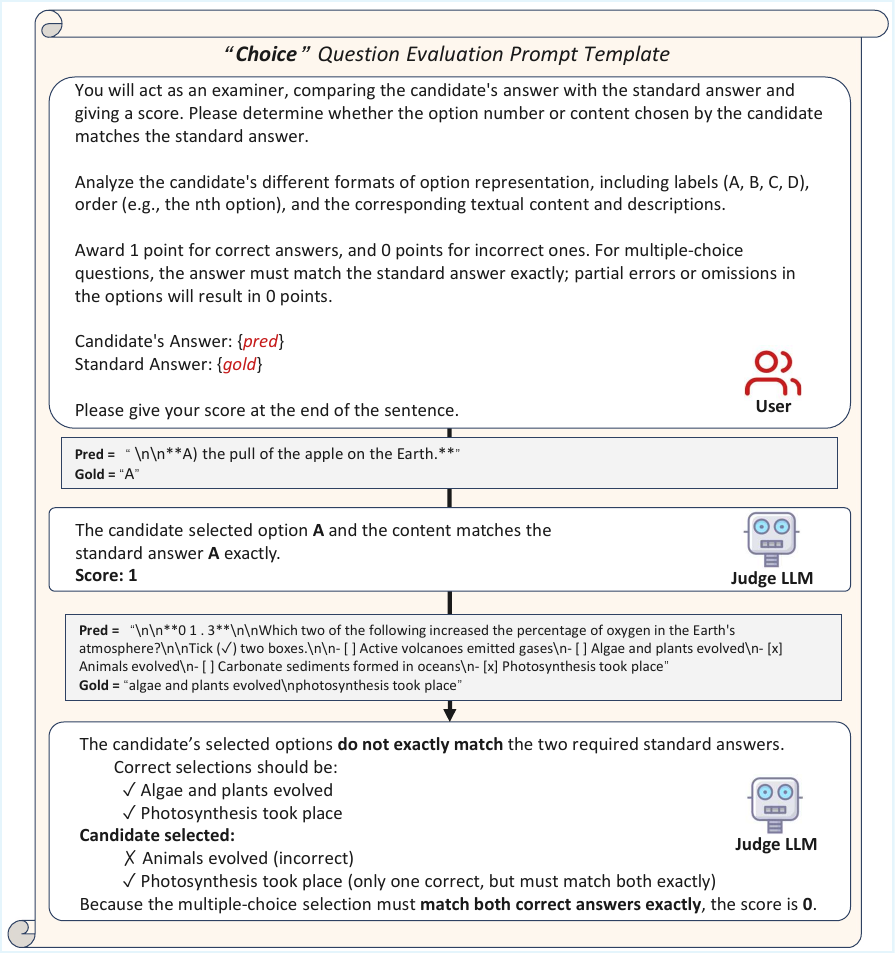}
        \caption{Scoring process for \textbf{Choice}-type questions.}
        \label{fig:choice}
    \end{subfigure}

    \caption{LLM-as-a-Judge scoring processes for Statement and
    Choice questions.}
    \label{fig:judge_statement_choice}
\end{figure}

\begin{figure}[p]
    \centering

    \begin{subfigure}[t]{0.485\linewidth}
        \centering
        \includegraphics[width=\linewidth]
        {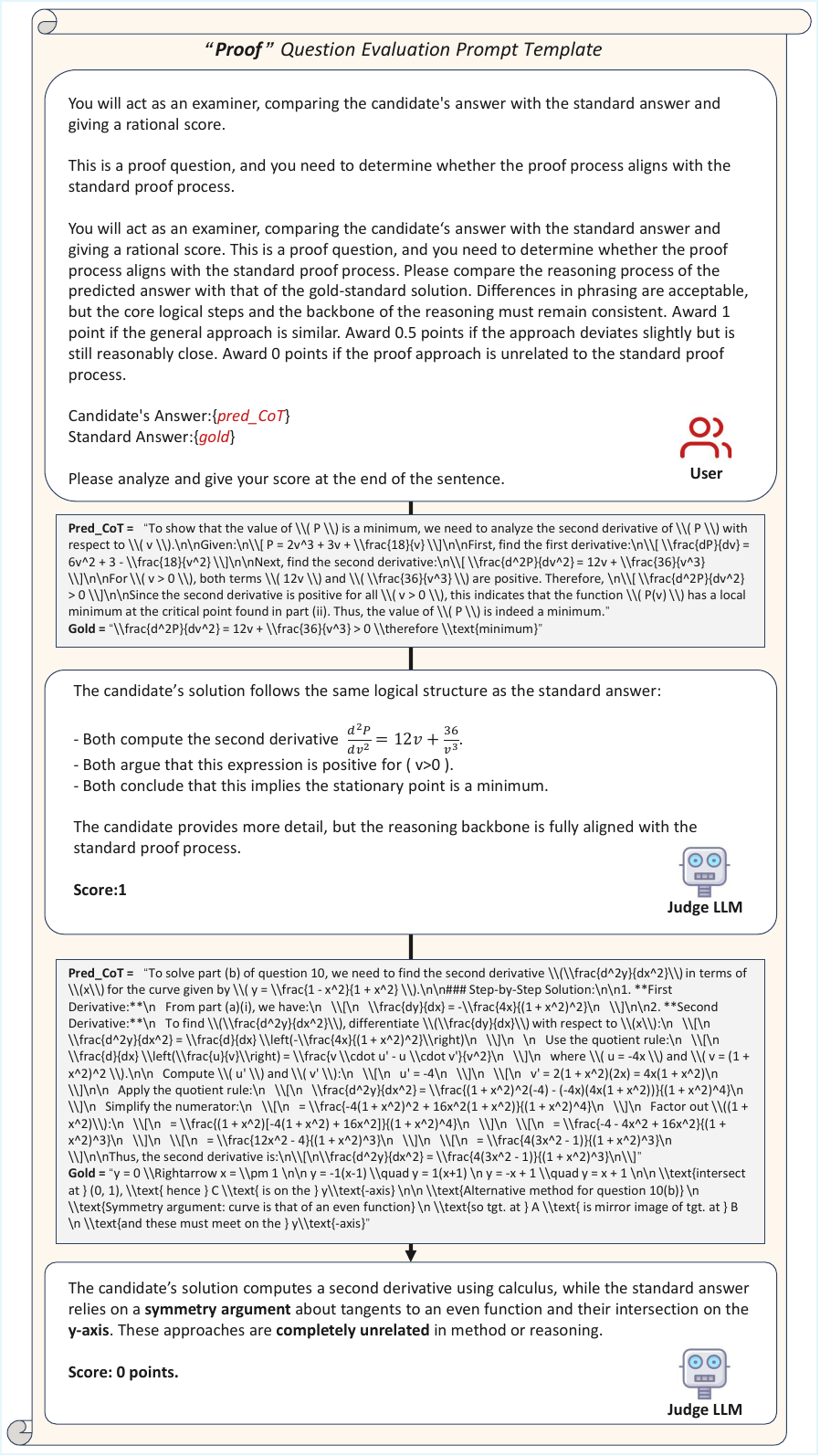}
        \caption{Scoring process for \textbf{Proof}-type questions.}
        \label{fig:proof}
    \end{subfigure}
    \hfill
    \begin{subfigure}[t]{0.485\linewidth}
        \centering
        \includegraphics[width=\linewidth]
        {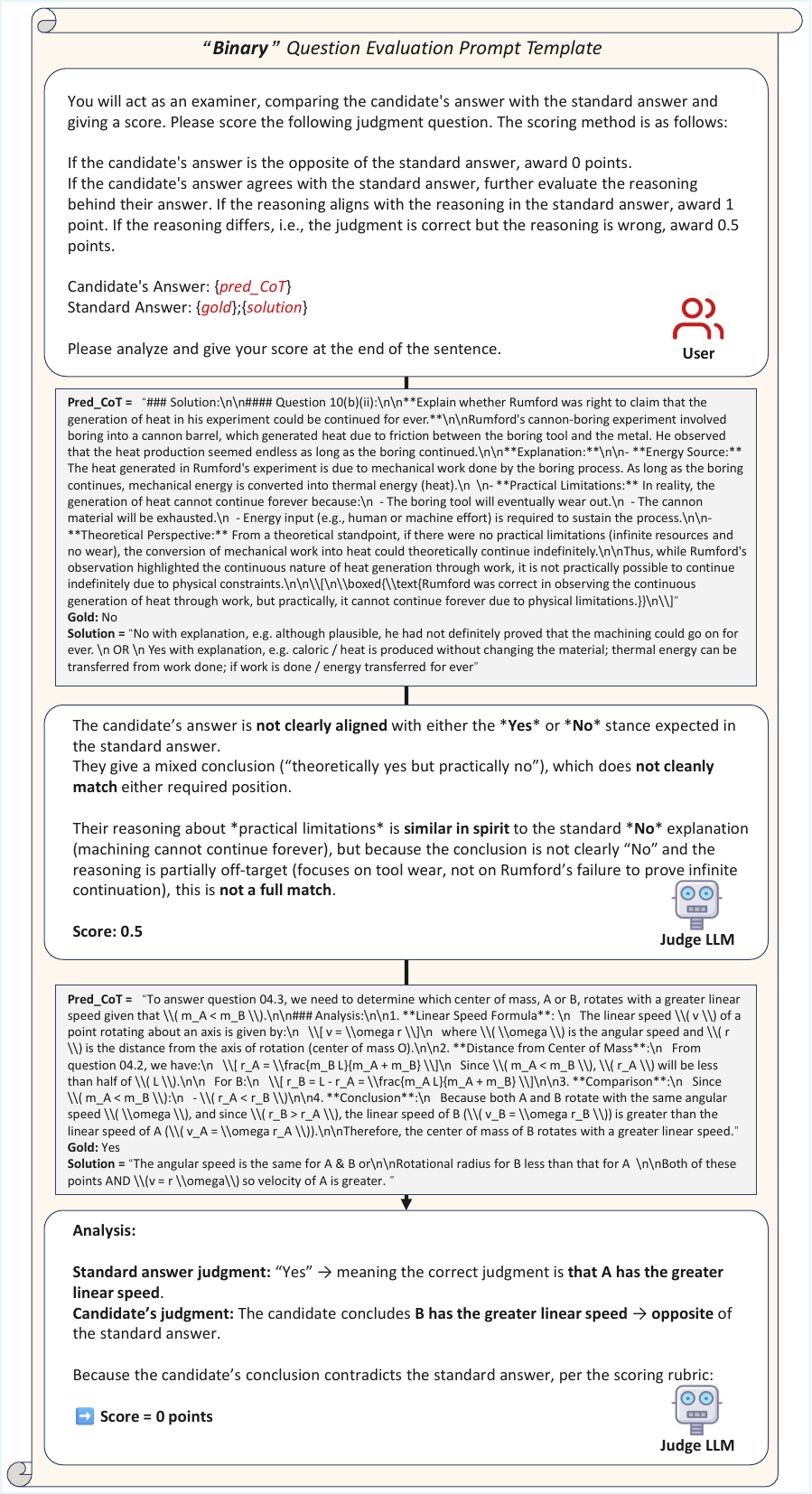}
        \caption{Scoring process for \textbf{Binary}-type questions.}
        \label{fig:binary}
    \end{subfigure}

    \caption{LLM-as-a-Judge scoring processes for Proof and
    Binary questions.}
    \label{fig:judge_proof_binary}
\end{figure}

The detailed content is shown in
Figures~\ref{fig:normal} and~\ref{fig:calculate},
Figures~\ref{fig:statement} and~\ref{fig:choice}, and
Figures~\ref{fig:proof} and~\ref{fig:binary}.

% Normal and Calculate

% Statement and Choice

% Proof and Binary

\end{document}